\documentclass[10pt,twocolumn,letterpaper]{article}

\usepackage{iccv}
\usepackage{times}
\usepackage{epsfig}
\usepackage{graphicx}
\usepackage{amsmath}
\usepackage{amssymb}
\usepackage{multirow}

\usepackage{amsfonts}
\usepackage[american]{babel}

\usepackage[pagebackref=true,breaklinks=true,letterpaper=true,colorlinks,urlcolor=black,bookmarks=false]{hyperref}

\iccvfinalcopy 

\def\iccvPaperID{3220} 

\ificcvfinal\fi
\begin{document}

\title{\ Image Inpainting with Learnable Bidirectional Attention Maps}

%
%
%
\author{\small{Chaohao Xie$^{1}$\footnotemark[2]\ \ , Shaohui Liu$^{1,3}$, Chao Li$^{2}$, Ming-Ming Cheng$^{4}$, Wangmeng Zuo$^{1,3}$\footnotemark[1]\ \ , Xiao Liu$^{2}$, Shilei Wen$^{2}$, Errui Ding$^{2}$} \\
	$^1$\small{Harbin Institute of Technology}, $^2$\small{Department of Computer Vision Technology (VIS), Baidu Inc.}\\
	$^3$\small{Peng Cheng Laboratory, Shenzhen}, $^4$\small{Nankai University}\\
	{\tt\small{viousxie@outlook.com}, } {\tt {\small\{\href{mailto:shliu@hit.edu.cn}{shliu}, \href{mailto:wmzuo@hit.edu.cn}{wmzuo}\}@hit.edu.cn}, } {\tt{\small{cmm@nankai.edu.cn}}} \\
	{\tt \small{\{\href{mailto:lichao40@baidu.com}{lichao40}, \href{mailto:liuxiao12@baidu.com}{liuxiao12}, \href{mailto:wenshilei@baidu.com}{wenshilei}, \href{mailto:dingerrui@baidu.com}{dingerrui}\}@baidu.com}}
}

\maketitle

\renewcommand{\thefootnote}{\fnsymbol{footnote}}
\footnotetext[2]{This work was done when Chaohao Xie was a research intern at Baidu}
\footnotetext[1]{Corresponding author}

\begin{abstract}
\vspace{-0.5em}
   %
   %
   Most convolutional network (CNN)-based inpainting methods adopt standard convolution to indistinguishably treat valid pixels and holes, making them limited in handling irregular holes and more likely to generate inpainting results with color discrepancy and blurriness.
   Partial convolution has been suggested to address this issue, but it adopts handcrafted feature re-normalization, and only considers forward mask-updating.
   In this paper, we present a learnable attention map module for learning feature re-normalization and mask-updating in an end-to-end manner, which is effective in adapting to irregular holes and propagation of convolution layers.
   Furthermore, learnable reverse attention maps are introduced to allow the decoder of U-Net to concentrate on filling in irregular holes instead of reconstructing both holes and known regions, resulting in our learnable bidirectional attention maps.
   Qualitative and quantitative experiments show that our method performs favorably against state-of-the-arts in generating sharper, more coherent and visually plausible inpainting results.
   The source code and pre-trained models will be available.

   %
   %
\end{abstract}

\vspace{-1.2em}
\section{Introduction}
\begin{figure*}[hbt]\label{fig:figure1}
\small
	\setlength{\tabcolsep}{2.0pt}
	\centering
	\begin{tabular}{ccccccc}
		\includegraphics[width=.135\textwidth]{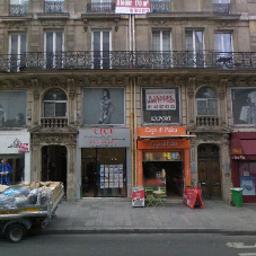}  &
		\includegraphics[width=.135\textwidth]{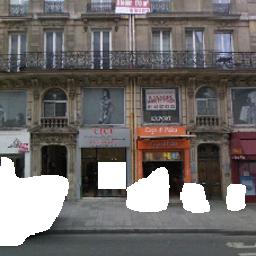}  &
		\includegraphics[width=.135\textwidth]{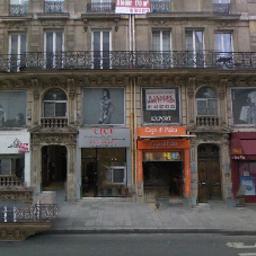}  &
		\includegraphics[width=.135\textwidth]{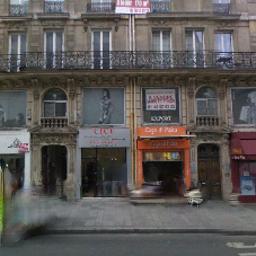}  &
		\includegraphics[width=.135\textwidth]{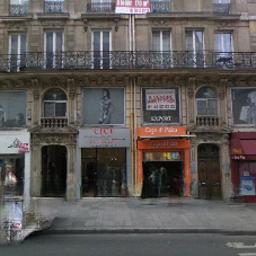}  &
		\includegraphics[width=.135\textwidth]{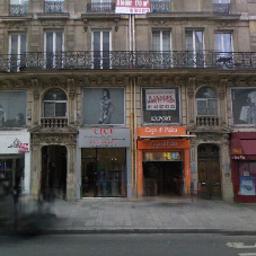}  &
		\includegraphics[width=.135\textwidth]{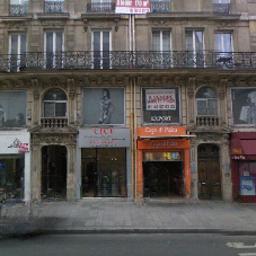}  \\
		
		\includegraphics[width=.135\textwidth]{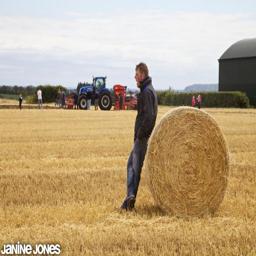}  &
		\includegraphics[width=.135\textwidth]{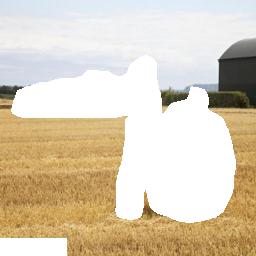}  &
		\includegraphics[width=.135\textwidth]{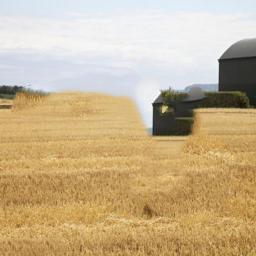}  &
		\includegraphics[width=.135\textwidth]{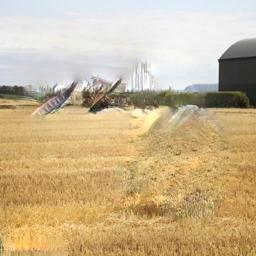}  &
		\includegraphics[width=.135\textwidth]{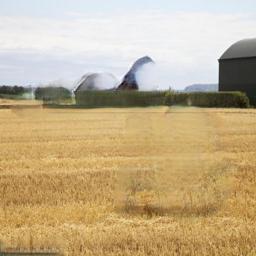}  &
		\includegraphics[width=.135\textwidth]{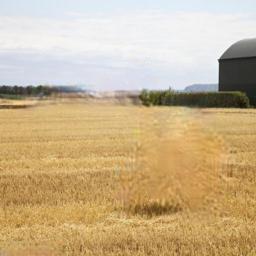}  &
		\includegraphics[width=.135\textwidth]{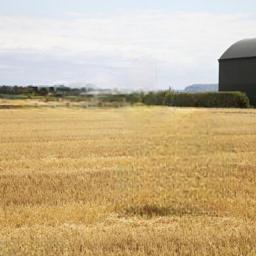}  \\

		(a) Original & (b) Input & (c) PM~\cite{Barnes:2009:PAR}  & (d) GL~\cite{IizukaGL} & (e) CA~\cite{yu2018generative} & (f) PConv~\cite{partialconv2017} & (g) Ours \\
		
	\end{tabular}
	\caption{Qualitative comparison of inpainting results by PatchMatch (PM)~\cite{Barnes:2009:PAR}, Global\&Local (GL)~\cite{IizukaGL},  Context Attention (CA)~\cite{yu2018generative}, and Partial Convolution (PConv)~\cite{partialconv2017}, and Ours.}
	\vspace{-1.6em}
\end{figure*}

\vspace{-0.4em}
Image inpainting~\cite{BertalmioInpainting}, aiming at filling in holes of an image, is a representative low level vision task with many real-world applications such as distracting object removal, occluded region completion, etc.
However, there may exist multiple potential solutions for the given holes in an image, \ie,
the holes can be filled with any plausible hypotheses coherent with the surrounding known regions.
And the holes can be of complex and irregular patterns, further increasing the difficulty of image inpainting.
Traditional exemplar-based methods~\cite{Barnes:2009:PAR,LeMeur2011examplar,XuPatchSparsity}, \eg, PatchMatch~\cite{Barnes:2009:PAR}, gradually fill in holes by searching and copying similar patches from known regions.
Albeit exemplar-based methods are effective in hallucinating detailed textures, they are still limited in capturing high-level semantics, and may fail to generate complex and non-repetitive structures (see Fig.~\ref{fig:figure1}(c)).



Recently, considerable progress has been made in applying deep convolutional networks (CNNs) to image inpainting~\cite{IizukaGL,pathakCVPR16context}.
Benefited from the powerful representation ability and large scale training, CNN-based methods are effective in hallucinating semantically plausible result.
And adversarial loss~\cite{Goodfellow_GAN} has also been deployed to improve the perceptual quality and naturalness of the result.
Nonetheless, most existing CNN-based methods usually adopt standard convolution which indistinguishably treats valid pixels and holes.
Thus, they are limited in handling irregular holes and more likely to generate inpainting results with color discrepancy and blurriness.
As a remedy, several post-processing techniques~\cite{IizukaGL,Yang_2017_CVPR} have been introduced but are still inadequate in resolving the artifacts (see Fig.~\ref{fig:figure1}(d)).


CNN-based methods have also been combined with exemplar-based one to explicitly incorporate the mask of holes for better structure recovery and detail enhancement~\cite{song_contextual_2018,Yan_2018_Shift,yu2018generative}.
In these methods, the mask is utilized to guide the propagation of the encoder features from known regions to the holes.
However, the copying and enhancing operation heavily increases the computational cost and is only deployed at one encoding and decoding layers.
As a result, they are better at filling in rectangular holes, and perform poorly on handling irregular holes (see Fig.~\ref{fig:figure1}(e)).



For better handling irregular holes and suppressing color discrepancy and blurriness, partial convolution (PConv)~\cite{partialconv2017} has been suggested.
In each PConv layer, mask convolution is used to make the output conditioned only on the unmasked input, and feature re-normalization is introduced for scaling the convolution output.
A mask-updating rule is further presented to update a mask for the next layer, making PConv very effective in handling irregular holes.
Nonetheless, PConv adopts hard 0-1 mask and handcrafted feature re-normalization by absolutely trusting all filling-in intermediate features.
Moreover, PConv considers only forward mask-updating and simply employs all-one mask for decoder features.


In this paper, we take a step forward and present the modules of learnable bidirectional attention maps for the re-normalization of features on both encoder and decoder of the U-Net~\cite{UNetRFB15a} architecture.
To begin with, we revisit PConv without bias, and show that the mask convolution can be safely avoided and the feature re-normalization can be interpreted as a re-normalization guided by hard 0-1 mask.
To overcome the limitations of hard 0-1 mask and handcrafted mask-updating, we present a learnable attention map module for learning feature re-normalization and mask-updating.
Benefited from the end-to-end training, the learnable attention map is effective in adapting to irregular holes and propagation of convolution layers.


Furthermore, PConv simply uses all-one mask on the decoder features, making the decoder should hallucinate both holes and known regions.
Note that the encoder features of known region will be concatenated, it is natural that the decoder is only required to focus on the inpainting of holes.
Therefore, we further introduce learnable reverse attention maps to allow the decoder of U-Net concentrate only on filling in holes, resulting in our learnable bidirectional attention maps.
In contrast to PConv, the deployment of learnable bidirectional attention maps empirically is beneficial to network training, making it feasible to include adversarial loss for improving visual quality of the result.

Qualitative and quantitative experiments are conducted on the Paris SteetView~\cite{doersch2015makes} and Places~\cite{zhou2017places} datasets to evaluate our proposed method.
The results show that our proposed method performs favorably against state-of-the-arts in generating sharper, more coherent and visually plausible inpainting results.
From Fig.~\ref{fig:figure1}(f)(g), our method is more effective in hallucinating clean semantic structure and realistic textures in comparison to PConv.
To sum up, the main contribution of this work is three-fold,
\begin{itemize}
\vspace{-0.8em}
 \item A learnable attention map module is presented for image inpainting. In contrast to PConv, the learnable attention maps are more effective in adapting to arbitrary irregular holes and propagation of convolution layers.
     \vspace{-0.8em}
 \item Forward and reverse attention maps are incorporated to constitute our learnable bidirectional attention maps, further benefiting the visual quality of the result.
     \vspace{-0.8em}
 \item Experiments on two datasets and real-world object removal show that our method performs favorably against state-of-the-arts in hallucinating shaper, more coherent and visually plausible results.
\end{itemize}



\vspace{-0.4em}
\section{Related Work}
\vspace{-0.4em}
In this section, we present a brief survey on the relevant work, especially the propagation process adopted in exemplar-based methods as well as the network architectures of CNN-based inpainting methods.


\begin{figure*}[ht]
\small
	\centering
	\includegraphics[width=\linewidth]{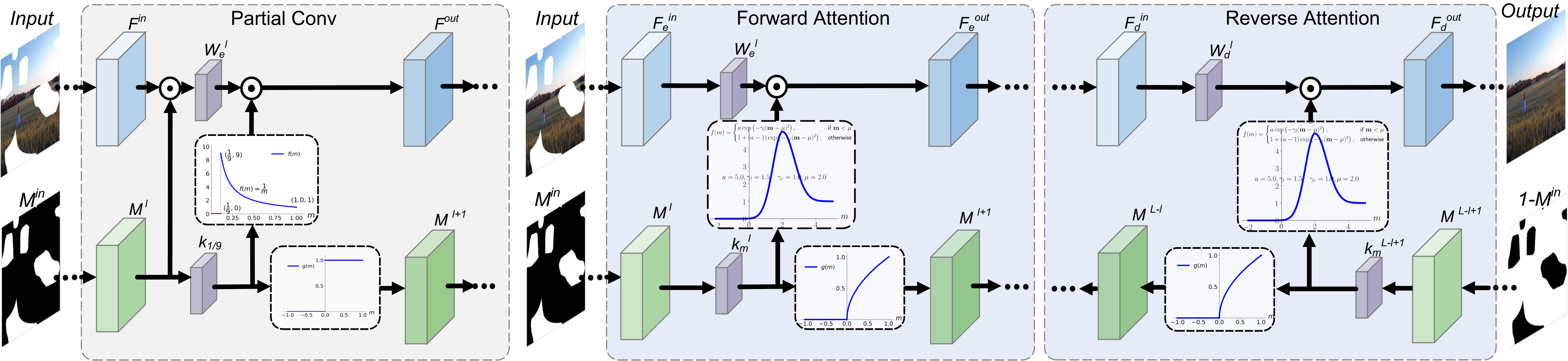}
	\begin{tabular}{lcl}
		\quad\quad\qquad(a) PConv \quad\quad\qquad\qquad & \quad\qquad (b) Learnable forward attention map & \quad(c) Learnable reverse attention map \\
	\end{tabular}
	\caption{Interplay models between mask and intermediate feature for PConv and our learnable bidirectional attention maps. Here, the white holes in $\mathbf{M}^{in}$ denotes missing region with value 0, and the black area denotes the known region with value 1.}
	\label{fig:figure2}
	\vspace{-1.6em}
\end{figure*}

\vspace{-0.4em}
\subsection{Exemplar-based Inpainting}
\vspace{-0.4em}
Most exemplar-based inpainting methods search and paste from the known regions to gradually fill in the holes from the exterior to the interior~\cite{Barnes:2009:PAR,Criminisi2004region,LeMeur2011examplar,XuPatchSparsity}, and their results highly depend on the propagation process.
In general, better inpainting result can be attained by first filling in structures and then other missing regions.
To guide the patch processing order, patch priority~\cite{KomodakisPriority,WilczkowiakBMVC05} measure has been introduced as the product of confidence term and data term.
While the confidence term is generally defined as the ratio of known pixels in the input patch, several forms of data terms have been proposed.
In particular, Criminisi \etal~\cite{Criminisi2004region} suggested a gradient-based data term for filling in linear structure with higher priority.
Xu and Sun~\cite{XuPatchSparsity} assumed that structural patches are sparsely distributed in an image, and presented a sparsity-based data term.
Le Meur~\etal~\cite{LeMeur2011examplar} adopted the eigenvalue discrepancy of structure tensor~\cite{DiZenzo1986gradient} as an indicator of structural patch.


\vspace{-0.4em}
\subsection{Deep CNN-based Inpainting}
\vspace{-0.4em}
Early CNN-based methods~\cite{KohlerSSHJR2014,RenShepardConv,XieDenoiseCNN} are suggested for handling images with small and thin holes.
In the past few years, deep CNNs have received upsurging interest and exhibited promising performance for filling in large holes.
Phatak \etal~\cite{pathakCVPR16context} adopted an encoder-decoder network (\ie, context-encoder), and incorporated reconstruction and adversarial losses for better recovering semantic structures.
%
%
Iizuka \etal~\cite{IizukaGL} combined both global and local discriminators for reproducing both semantically plausible structures and locally realistic details.
Wang \etal~\cite{WangMulticolumn} suggested a generative multi-column CNN incorporating with confidence-driven reconstruction loss and implicit diversified MRF (ID-MRF) term.

Multi-stage methods have also been investigated to ease the difficulty of training deep inpainting networks.
Zhang \etal~\cite{ZhangPGN} presented a progressive generative networks~(PGN) for filling in holes with multiple phases, while LSTM is deployed to exploit the dependencies across phases.
Nazeri \etal~\cite{nazeri2019edgeconnect} proposed a two-stage model EdgeConnect first predicting salient edges and then generating inpainting result guided by edges.
Instead, Xiong \etal~\cite{Xiong_2019_CVPR} presented foreground-aware inpainting, which involves three stages, \ie, contour detection, contour completion and image completion, for the disentanglement of structure inference and content hallucination.

In order to combine exemplar-based and CNN-based methods, Yang \etal~\cite{Yang_2017_CVPR} suggested multi-scale neural patch synthesis (MNPS) to refine the result of context-encoder via joint optimization with the holistic content and local texture constraints.
Other two-stage feed-forward models, \eg, contextual attention~\cite{song_contextual_2018} and patch-swap~\cite{yu2018generative}, are further developed to overcome the high computational cost of MNPS while explicitly exploiting image features of known regions.
Concurrently, Yan \etal~\cite{Yan_2018_Shift} modified the U-Net to form an one-stage network, \ie, Shift-Net, to utilize the shift of encoder feature from known regions for better reproducing plausible semantics and detailed contents.
Most recently, Zheng \etal~\cite{Zheng2019Pluralistic} introduced an enhanced short+long term attention layer, and presented a probabilistic framework with two parallel paths for pluralistic inpainting.


Most existing CNN-based inpainting methods are usually not well suited for handling irregular holes.
To address this issue, Liu \etal~\cite{partialconv2017} proposed a partial convolution (PConv) layer involving three steps, \ie, mask convolution, feature re-normalization, and mask-updating.
Yu \etal~\cite{yu2018free} provided gated convolution which learns channel-wise soft mask by considering both corrupted images, masks and user sketches.
However, PConv adopts handcrafted feature re-normalization and only considers forward mask-updating, making it still limited in handling color discrepancy and blurriness (see Fig.~\ref{fig:figure1}(d)).

\vspace{-0.4em}
\section{Proposed Method}
\vspace{-0.4em}
In this section, we first revisit PConv, and then present our learnable bidirectional attention maps.
Subsequently, the network architecture and learning objective of our method are also provided.

\begin{figure*}[ht]
	\centering
	\includegraphics[width=\linewidth]{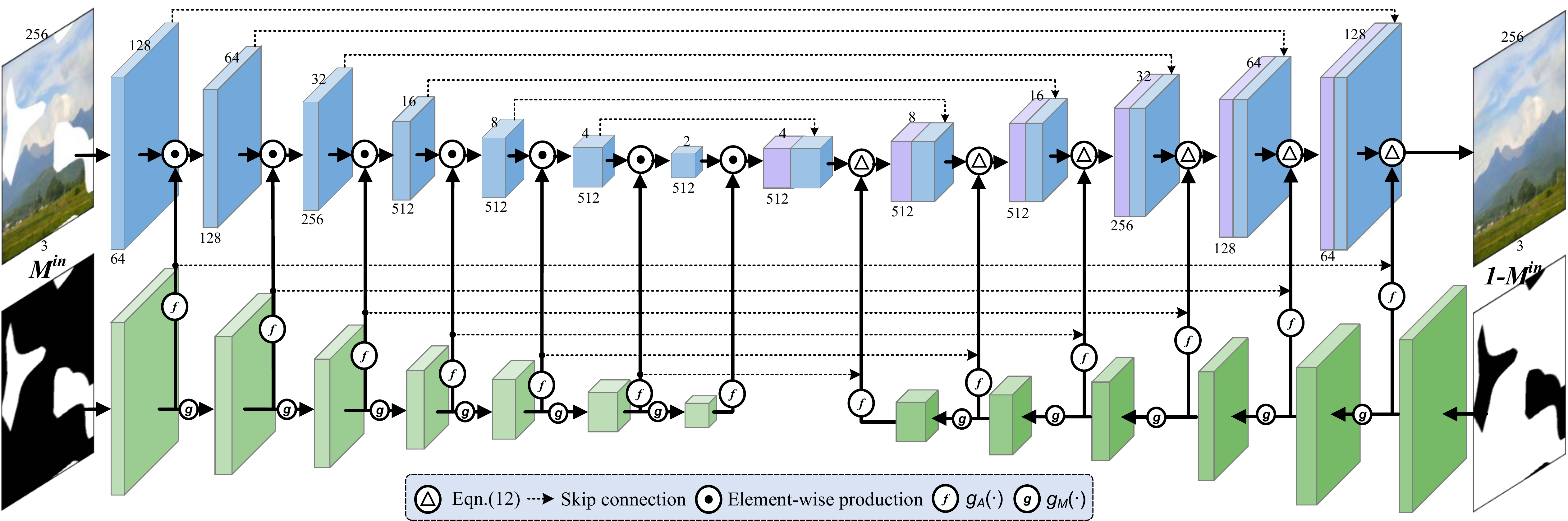}
	\caption{The network architecture of our model. The circle with triangle inside denotes operation form of Eqn.(~\ref{PConv-2-r}), $g_A$ and $g_M$ represent activation functions of Eqn.(~\ref{Activation_gA}) and mask updating function of Eqn.(~\ref{Activation_gM}).}
	\label{model}
	\vspace{-1.6em}
\end{figure*}

\vspace{-0.4em}
\subsection{Revisiting Partial Convolution}
\vspace{-0.4em}
A PConv~\cite{partialconv2017} layer generally involves three steps, \ie, (i) mask convolution, (ii) feature re-normalization, and (iii) mask-updating.
Denote by $\mathbf{F}^{in}$ the input feature map and $\mathbf{M}$ the corresponding hard 0-1 mask.
We further let $\mathbf{W}$ be the convolution filter and $b$ be its bias.
To begin with, we introduce the convolved mask $\mathbf{M}^c = \mathbf{M} \otimes \mathbf{k}_{\frac{1}{9}}$, where $\otimes$ denotes the convolution operator, $\mathbf{k}_{\frac{1}{9}}$ denotes a $3 \times 3$ convolution filter with each element $\frac{1}{9}$.
The process of PConv can be formulated as,
\begin{equation}\label{PConv-1}
\small
\mbox{(i) } \mathbf{F}^{conv} =
\mathbf{W}^{T}(\mathbf{F}^{in} \odot \mathbf{M}),
\end{equation}
\begin{equation}\label{PConv-2}
\small
\mbox{(ii) } \mathbf{F}^{out} =
\begin{cases}
\mathbf{F}^{conv} \odot f_A(\mathbf{M}^c) + b, & \mbox{if}\ \mathbf{M}^c > 0 \\
0, & \mbox{otherwise}
\end{cases}
\end{equation}
\begin{equation}\label{PConv-3}
\small
\mbox{(iii) }  \mathbf{M}^{\prime} = f_M(\mathbf{M}^c)
\end{equation}
where $\mathbf{A} = f_A(\mathbf{M}^c)$ denotes the attention map, and $\mathbf{M}^{\prime} = f_M(\mathbf{M}^c)$ denotes the updated mask.
We further define the activation functions for attention map and updated mask as,
\begin{equation}\label{Activation_A}
\small
f_A(\mathbf{M}^c) =
\begin{cases}
\frac{1}{\mathbf{M}^c}, & \mbox{if} \ \mathbf{M}^c > 0 \\
0, & \mbox{otherwise}
\end{cases}
\end{equation}
\begin{equation}\label{Activation_M}
\small
f_M(\mathbf{M}^c) =
\begin{cases}
1, & \mbox{if} \ \mathbf{M}^c > 0 \\
0, & \mbox{otherwise}
\end{cases}
\end{equation}

From Eqns.~(\ref{PConv-1})$\sim$(\ref{Activation_M}) and Fig.~\ref{fig:figure2}(a), PConv can also be explained as a special interplay model between mask and convolution feature map.
However, PConv adopts the handcrafted convolution filter $\mathbf{k}_{\frac{1}{9}}$ as well as handcrafted activation functions $f_A(\mathbf{M}^c)$ and $f_M(\mathbf{M}^c)$, thereby giving some leeway for further improvements.
Moreover, the non-differential property of $f_M(\mathbf{M}^c)$ also increases the difficulty of end-to-end learning.
To our best knowledge, it remains a difficult issue to incorporate adversarial loss to train a U-Net with PConv.
Furthermore, PConv only considers the mask and its updating for encoder features.
As for decoder features, it simply adopts all-one mask, making PConv limited in filling holes.

\vspace{-0.4em}
\subsection{Learnable Attention Maps}
\vspace{-0.4em}

The convolution layer without bias has been widely adopted in U-Net for image-to-image translation~\cite{isola2017cvpr} and image inpainting~\cite{Yan_2018_Shift}.
When the bias is removed, it can be readily seen from Eqn.~(\ref{PConv-2}) that the convolution features in updated holes are zeros.
Thus, the mask convolution in Eqn.~(\ref{PConv-1}) is equivalently rewritten as standard convolution,
\begin{equation}\label{PConv-1-nobias}
\small
\mbox{(i) } \mathbf{F}^{conv} =
\mathbf{W}^{T}\mathbf{F}^{in}.
\end{equation}
Then, the feature re-normalization in Eqn.~(\ref{PConv-2}) can be interpreted as the element-wise product of convolution feature and attention map,
\begin{equation}\label{PConv-2-nobias}
\small
\mbox{(ii) } \mathbf{F}^{out} = \mathbf{F}^{conv} \odot f_A(\mathbf{M}^c).
\end{equation}
%
%
Even though, the handcrafted convolution filter $\mathbf{k}_{\frac{1}{9}}$ is fixed and not adapted to the mask.
The activation function for updated mask absolutely trusts the inpainting result in the region $\mathbf{M}^c > 0$, but it is more sensible to assign higher confidence to the region with higher $\mathbf{M}^c$.

To overcome the above limitations, we suggest learnable attention map which generalizes PConv without bias from three aspects.
First, to make the mask adaptive to irregular holes and propagation along with layers, we substitute $\mathbf{k}_{\frac{1}{9}}$ with layer-wise and learnable convolution filters $\mathbf{k}_{\mathbf{M}}$.
Second, instead of hard 0-1 mask-updating, we modify the activation function for updated mask as,
\begin{equation}\label{Activation_gM}
\small
g_M(\mathbf{M}^c) = \left(ReLU(\mathbf{M}^c)\right)^{\alpha},
\end{equation}
where $\alpha \geq 0$ is a hyperparameter and we set $\alpha = 0.8$.
One can see that $g_M(\mathbf{M}^c)$ degenerates into $f_M(\mathbf{M}^c)$ when $\alpha = 0$.
Third, we introduce an asymmetric Gaussian-shaped form as the activation function for attention map,
\begin{equation}\label{Activation_gA}
\small
g_A(\mathbf{M}^c) \!\!=\!\!
\begin{cases}
a \exp\left( - {\gamma_l (\mathbf{M}^c - \mu)^2} \right), & \mbox{if} \ \mathbf{M}^c \!<\! \mu \\
1 \!+\! (a\!-\!1) \exp\left( - {\gamma_r (\mathbf{M}^c \!-\! \mu)^2} \right), & \mbox{else}
\end{cases}
\end{equation}
where $a$, $\mu$, $\gamma_l$, and $\gamma_r$ are the learnable parameters, we initialize them as $a = 1.1, \mu=2.0,\gamma_l = 1.0, \gamma_r = 1.0$ and learn them in an end-to-end manner.

To sum up, the learnable attention map adopt Eqn.~(\ref{PConv-1-nobias}) in Step (i),
and the next two steps are formulated as,
\begin{equation}\label{PConv-forward-2}
\small
\mbox{(ii) } \mathbf{F}^{out} = \mathbf{F}^{conv} \odot g_A(\mathbf{M}^c),
\end{equation}
\begin{equation}\label{PConv-forward-3}
\small
\mbox{(iii) }  \mathbf{M}^{\prime} = g_M(\mathbf{M}^c).
\end{equation}
Fig.~\ref{fig:figure2}(b) illustrates the interplay model of learnable attention map.
In contrast to PConv, our learnable attention map is more flexible and can be end-to-end trained, making it effective in adapting to irregular holes and propagation of convolution layers.

\begin{figure*}[hbt]
	\small
	\setlength{\tabcolsep}{2.0pt}
	\centering
	\begin{tabular}{ccccccc}
		\includegraphics[width=.135\textwidth]{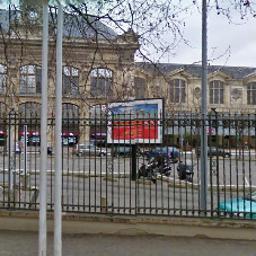}  &
		\includegraphics[width=.135\textwidth]{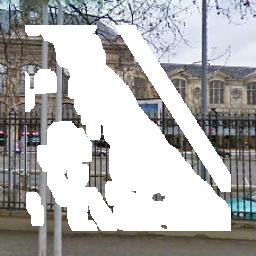}  &
		\includegraphics[width=.135\textwidth]{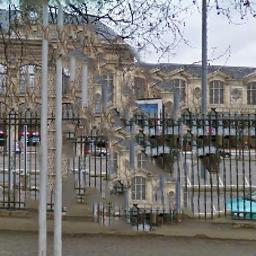}  &
		\includegraphics[width=.135\textwidth]{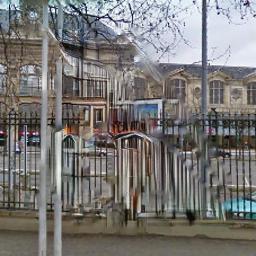}  &
		\includegraphics[width=.135\textwidth]{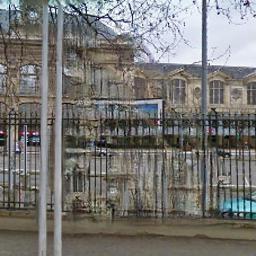}  &
		\includegraphics[width=.135\textwidth]{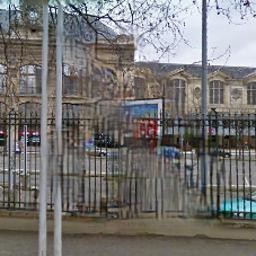}  &
		\includegraphics[width=.135\textwidth]{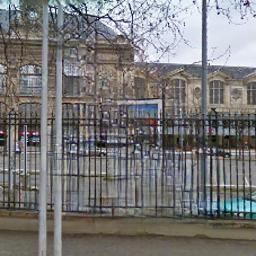}  \\
		
		\includegraphics[width=.135\textwidth]{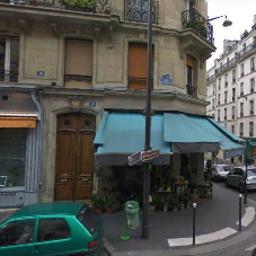}  &
		\includegraphics[width=.135\textwidth]{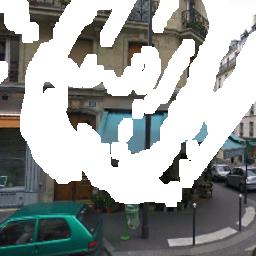}  &
		\includegraphics[width=.135\textwidth]{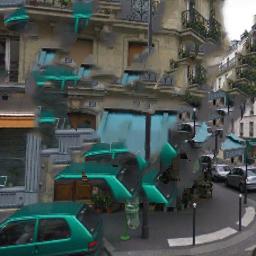}  &
		\includegraphics[width=.135\textwidth]{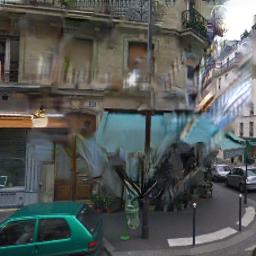}  &
		\includegraphics[width=.135\textwidth]{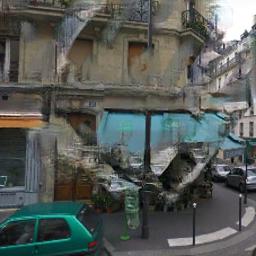}  &
		\includegraphics[width=.135\textwidth]{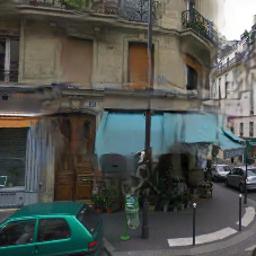}  &
		\includegraphics[width=.135\textwidth]{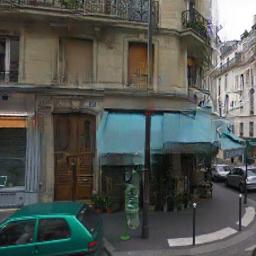}  \\
		
		\includegraphics[width=.135\textwidth]{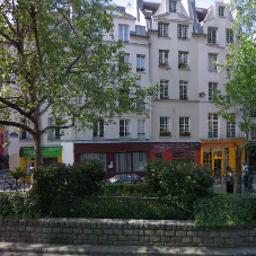}  &
		\includegraphics[width=.135\textwidth]{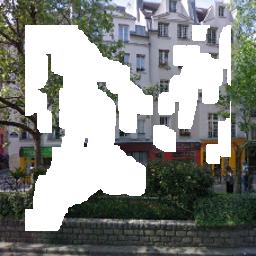}  &
		\includegraphics[width=.135\textwidth]{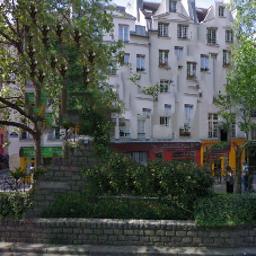}  &
		\includegraphics[width=.135\textwidth]{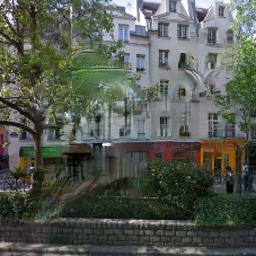}  &
		\includegraphics[width=.135\textwidth]{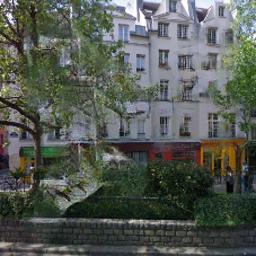}  &
		\includegraphics[width=.135\textwidth]{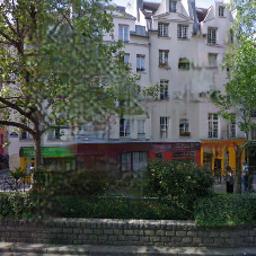}  &
		\includegraphics[width=.135\textwidth]{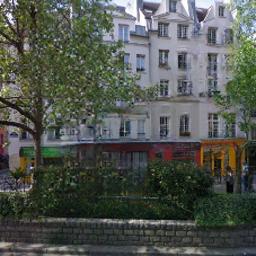}  \\
		
		Original & Input & PM~\cite{Barnes:2009:PAR} & GL~\cite{IizukaGL} & CA~\cite{yu2018generative} & PConv~\cite{partialconv2017} & Ours \\
		
	\end{tabular}
	\caption{Qualitative comparison on Paris StreetView dataset. Comparison with PatchMatch (PM)~\cite{Barnes:2009:PAR},  Global\&Local(GL)~\cite{IizukaGL}, Context Attention(CA)~\cite{yu2018generative}, PConv~\cite{partialconv2017} and Ours.}
	\label{fig:paris}
	\vspace{-1.64em}
\end{figure*}

\vspace{-0.4em}
\subsection{Learnable Bidirectional Attention Maps}
\vspace{-0.4em}

When incorporating PConv with U-Net for inpainting, the method~\cite{partialconv2017} only updates the masks along with the convolution layers for encoder features.
%
%
However, all-one mask is generally adopted for decoder features.
As a result, the $(L-l)$-th layer of decoder feature in both known regions and holes should be hallucinated using both $(l+1)$-th layer of encoder feature and $(L-l-1)$-th layer of decoder feature.
Actually, the $l$-th layer of encoder feature will be concatenated with the $(L-l)$-th layer of decoder feature, and we can only focus on the generation of the $(L-l)$-th layer of decoder feature in the holes.

\noindent We further introduce learnable reverse attention maps to the decoder features.
Denote by $\mathbf{M}^c_e$ the convolved mask for encoder feature $\mathbf{F}^{in}_e$.
Let $\mathbf{M}^c_d = \mathbf{M}_d \otimes \mathbf{k}_{\mathbf{M}_d}$ be the convolved mask for decoder feature $\mathbf{F}^{in}_d$.
The first two steps of learnable reverse attention map can be formulated as,
\begin{equation}\label{PConv-2-r}
\small
\mbox{(i\&ii) } \mathbf{F}^{out}_d \!=\! (\mathbf{W}_e^T \mathbf{F}^{in}_e) \odot g_A(\mathbf{M}^c_e) \!+\! (\mathbf{W}_d^T \mathbf{F}^{in}_d) \!\odot\! g_A(\mathbf{M}^c_d).
\end{equation}
where $\mathbf{W}_e$ and $\mathbf{W}_d$ are the convolution filters. And we define $g_A(\mathbf{M}^c_d)$ as the reverse attention map.
Then, the mask $\mathbf{M}^c_d$ is updated and deployed to the former decoder layer,
\begin{equation}\label{PConv-3-r}
\small
\mbox{(iii) }  \mathbf{M}^{\prime}_d = g_M(\mathbf{M}^c_d).
\end{equation}
Fig.~\ref{fig:figure2}(c) illustrates the interplay model of reverse attention map.
In contrast to forward attention maps, both encoder feature (mask) and decoder feature (mask) are considered.
Moreover, the updated mask in reverse attention map is applied to the former decoder layer, while that in forward attention map is applied to the next encoder layer.



By incorporating forward and reverse attention maps with U-Net, Fig.~\ref{model} shows the full learnable bidirectional attention maps.
Given an input image $I^{in}$ with irregular holes, we use $\mathbf{M}^{in}$ to denote the binary mask, where ones indicate the valid pixels and zeros indicate the pixels in holes.
%
%
From Fig.~\ref{model}, the forward attention maps take $\mathbf{M}^{in}$ as the input mask for the re-normalization of the first layer of encoder feature, and gradually update and apply the mask to next encoder layer.
In contrast, the reverse attention maps take $1 - \mathbf{M}^{in}$ as the input for the re-normalization of the last (\ie, $L$-th) layer of decoder feature, and gradually update and apply the mask to former decoder layer.
Benefited from the end-to-end learning, our learnable bidirectional attention maps (LBAM) are more effective in handling irregular holes.
The introduction of reverse attention maps allows the decoder concentrate only on filling in irregular holes, which is also helpful to inpainting performance.
Our LBAM is also beneficial to network training, making it feasible to exploit adversarial loss for improving visual quality.

\begin{figure*}[hbt]
\small
	\setlength{\tabcolsep}{2.0pt}
	\centering
	\begin{tabular}{ccccccc}
		\includegraphics[width=.135\textwidth]{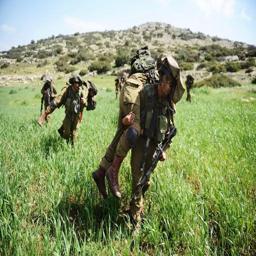}  &
		\includegraphics[width=.135\textwidth]{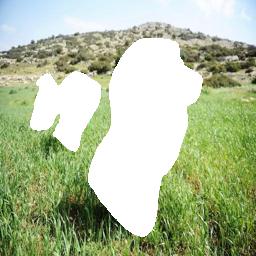}  &
		\includegraphics[width=.135\textwidth]{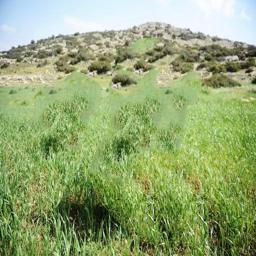}  &
		\includegraphics[width=.135\textwidth]{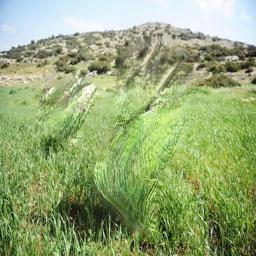}  &
		\includegraphics[width=.135\textwidth]{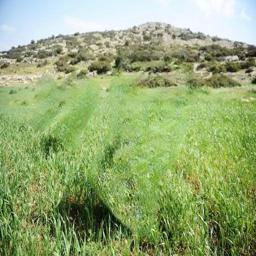}  &
		\includegraphics[width=.135\textwidth]{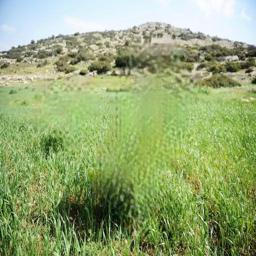}  &
		\includegraphics[width=.135\textwidth]{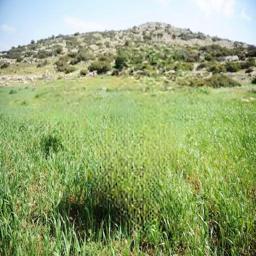}  \\
		
		\includegraphics[width=.135\textwidth]{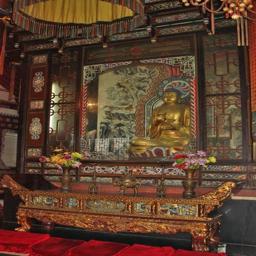}  &
		\includegraphics[width=.135\textwidth]{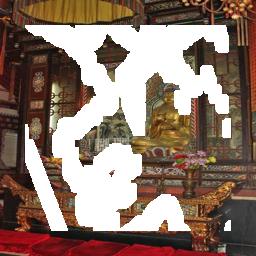}  &
		\includegraphics[width=.135\textwidth]{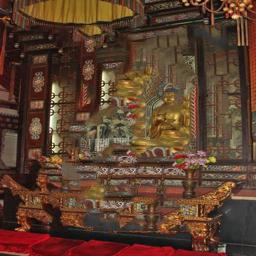}  &
		\includegraphics[width=.135\textwidth]{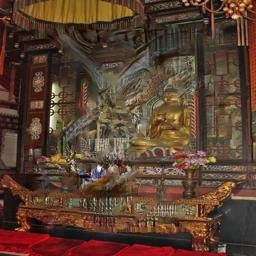}  &
		\includegraphics[width=.135\textwidth]{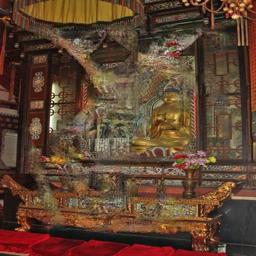}  &
		\includegraphics[width=.135\textwidth]{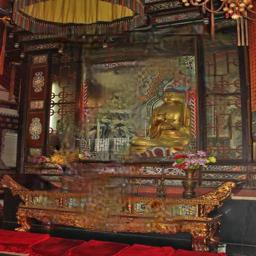}  &
		\includegraphics[width=.135\textwidth]{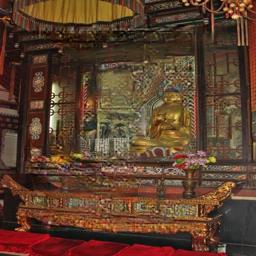}  \\
		
		\includegraphics[width=.135\textwidth]{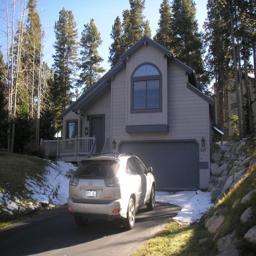}  &
		\includegraphics[width=.135\textwidth]{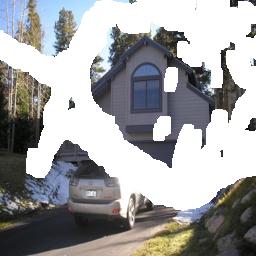}  &
		\includegraphics[width=.135\textwidth]{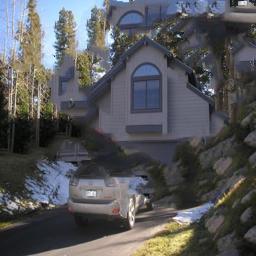}  &
		\includegraphics[width=.135\textwidth]{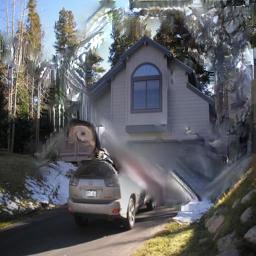}  &
		\includegraphics[width=.135\textwidth]{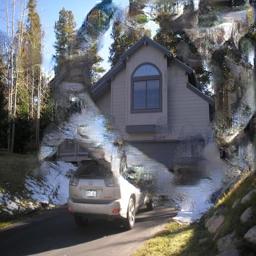}  &
		\includegraphics[width=.135\textwidth]{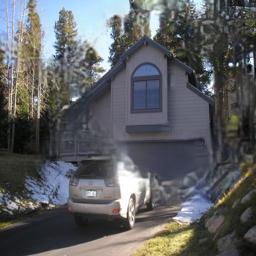}  &
		\includegraphics[width=.135\textwidth]{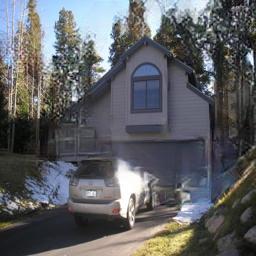}  \\
		
		Original & Input & PM~\cite{Barnes:2009:PAR}  & GL~\cite{IizukaGL} & CA~\cite{yu2018generative} & PConv~\cite{partialconv2017} & Ours \\
		
	\end{tabular}
	\caption{Qualitative comparison on Places dataset. Comparison with PatchMatch (PM)~\cite{Barnes:2009:PAR},  Global\&Local(GL)~\cite{IizukaGL}, Context Attention(CA)~\cite{yu2018generative}, PConv~\cite{partialconv2017} and Ours.}
	\label{fig:places}
	\vspace{-1.5em}
\end{figure*}

\vspace{-0.4em}
\subsection{Model Architecture}
\vspace{-0.4em}
We modify the U-Net architecture~\cite{isola2017cvpr} of 14 layers by removing the bottleneck layer and incorporating with bidirectional attention maps (see Fig.~\ref{model}).
In particular, forward attention layers are applied to the first six layers of encoder, while reverse attention layers are adopted to the last six layers of decoder.
%
%
For all the U-Net layers and the forward and reverse attention layers, we use convolution filters with the kernel size of $4 \times 4$, stride $2$ and padding $1$, and no bias parameters are used.
In the U-Net backbone, batch normalization and leaky ReLU nonlinearity are used to the features after re-normalization, and tanh nonlinearity is deployed right after
convolution for the last layer.
%
%
Fig.~\ref{model} also provides the size of feature map for each layer, and more details of the network architecture are given in the suppl.

\vspace{-0.4em}
\subsection{Loss Functions}
\vspace{-0.4em}
For better recovery of texture details and semantics, we incorporate pixel reconstruction loss, perceptual loss~\cite{Johnson2016Perceptual}, style loss~\cite{Gatys2016ImageST} and adversarial loss~\cite{Goodfellow_GAN} to train our LBAM.

\vspace{0.2em}
\noindent\textbf{Pixel Reconstruction Loss}. Denote by $I^{in}$ the input image with holes, $\mathbf{M}^{in}$ the binary mask region, and $I^{gt}$ the ground-truth image.
The output of our LBAM can be defined as $I^{out}= \Phi(I^{in}, \mathbf{M}^{in}; {\Theta})$, where ${\Theta}$ denotes the model parameters to be learned.
We adopt the $\ell_{1}$-norm error of the output image as the pixel reconstruction loss,
\vspace{-0.4em}
\begin{equation}
\vspace{-0.2em}
\small
\label{l1loss}
\mathcal{L}_{\ell_{1}} = {\parallel I^{out} - I^{gt} \parallel}_{1}.
\end{equation}

\vspace{0.1em}
\noindent\textbf{Perceptual Loss}.
The $\ell_{1}$-norm loss is limited in capturing high-level semantics and is not consistent with the human perception of image quality.
To alleviate this issue, we introduce the perceptual loss $\mathcal{L}_{perc}$ defined on the VGG-16 network~\cite{SimonyanZ14a} pre-trained on ImageNet~\cite{ILSVRC15},
\vspace{-0.4em}
\begin{equation}\label{perc-loss}
\small
\vspace{-0.2em}
\mathcal{L}_{perc} = \frac{1}{N}\sum\nolimits_{i=1}^{N} \parallel \mathcal{P}^{i}(I^{gt}) - \mathcal{P}^{i}(I^{out})  \parallel^2
\end{equation}
where $\mathcal{P}^{i}(\cdot)$ is the feature maps of the $i$-th pooling layer. In our implementation, we use $pool$-1, $pool$-2, and $pool$-3 layers of the pre-trained VGG-16.

\vspace{0.1em}
\noindent\textbf{Style Loss.}\ \ For better recovery of detailed textures, we further adopt the style loss defined on the feature maps from the pooling layers of VGG-16.
Analogous to~\cite{partialconv2017}, we construct a Gram matrix from each layer of feature map.
Suppose that the size of feature map $\mathcal{P}^{i}(I)$ is $H_{i} \times W_{i} \times C_{i}$.
The style loss can then be defined as,
\vspace{-0.4em}
\begin{equation}\label{style-loss}\small
\begin{split}
\mathcal{L}_{style} & = \frac{1}{N} \sum\nolimits_{i=1}^{N} \frac{1}{C_{i} \times C_{i}}  \times  \\
&\parallel \mathcal{P}^{i}(I^{gt}) (\mathcal{P}^{i}(I^{gt}))^{T} - \mathcal{P}^{i}(I^{out}) (\mathcal{P}^{i}(I^{out}))^{T} \parallel^2
\end{split}
\end{equation}

\begin{figure*}[hbt]
\small
	\setlength{\tabcolsep}{2.0pt}
	\centering
	\begin{tabular}{ccccc}
		\includegraphics[width=.192\textwidth]{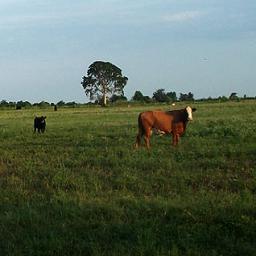}  &
		\includegraphics[width=.192\textwidth]{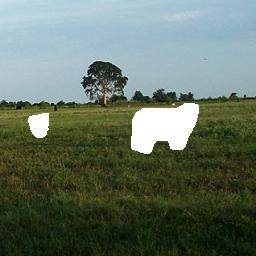}  &
		\includegraphics[width=.192\textwidth]{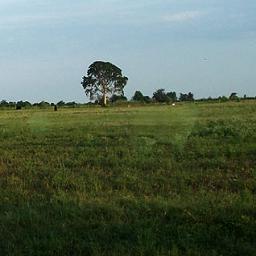}  &
		\includegraphics[width=.192\textwidth]{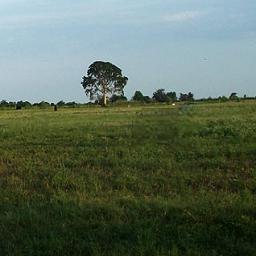}  &
		\includegraphics[width=.192\textwidth]{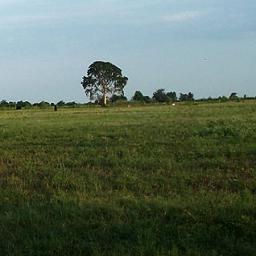}  \\
		
		\includegraphics[width=.192\textwidth]{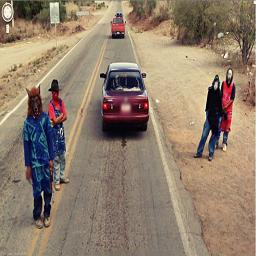}  &
		\includegraphics[width=.192\textwidth]{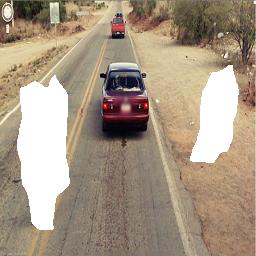}  &
		\includegraphics[width=.192\textwidth]{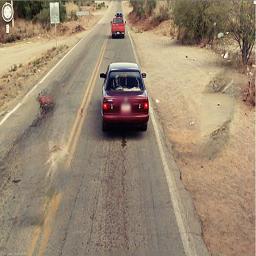}  &
		\includegraphics[width=.192\textwidth]{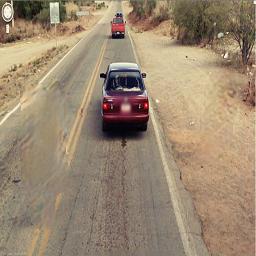}  &
		\includegraphics[width=.192\textwidth]{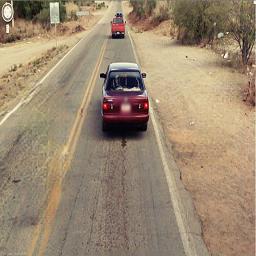}  \\
		
		Original & Input & CA~\cite{yu2018generative} & PConv~\cite{partialconv2017} & Ours \\
		
	\end{tabular}
	\caption{Results on real-world images. From left to right are: original image, input with objects masked (white area), Context Attention (CA)~\cite{yu2018generative}, PConv~\cite{partialconv2017}, and Ours.}
	\label{fig:real}
	\vspace{-1.64em}
\end{figure*}

\vspace{0.1em}
\noindent\textbf{Adversarial Loss}.
Adversarial loss~\cite{Goodfellow_GAN} has been widely adopted in image generation~\cite{salimans2016improved,pixelCNN,han2017stackgan} and low level vision~\cite{Ledig2016a} for improving the visual quality of generated images.
%
%
%
In order to improve the training stability of GAN, Arjovsky \etal~\cite{MartinWGAN} exploit the Wasserstein distance
for measuring the distribution discrepancy between generated and real images, and Gulrajani \etal~\cite{ishaan2017improved} further introduce gradient penalty for enforcing the Lipschitz constraint in discriminator.
Following~\cite{ishaan2017improved}, we formulate the adversarial loss as,
\vspace{-0.4em}
\begin{equation}\label{adv-loss}
\vspace{-0.2em}
\small
\begin{split}
\mathcal{L}_{adv} &= \min_{\Theta} \max_{D} E_{I^{gt} \sim p_{data}(I^{gt})} D(I^{gt}) \\
& - E_{I^{out} \sim p_{data}(I^{out})} D(I^{out})\\
& + \lambda E_{\hat{I} \sim p_{\hat{I}}}((\parallel \nabla_{\hat{I}}D(\hat{I}) \parallel)^2 - 1)^2
\end{split}
\end{equation}
where $D(\cdot)$ represents the discriminator. $\hat{I}$ is sampled from $I^{gt}$ and $I^{out}$ by linear interpolation with a randomly selected factor, $\lambda$ is set to 10 in our experiments.
We empirically find that it is difficult to train the PConv model when including adversarial loss.
Fortunately, the incorporation of learnable attention maps is helpful to ease the training, making it feasible to learn LBAM with adversarial loss.
Please refer to the suppl. for the network architecture of the 7-layer discriminator used in our implementation.

\noindent\textbf{Model Objective}\ \ Taking the above loss functions into account, the model objective of our LBAM can be formed as,
\begin{equation}\label{joint-loss}
\small
\mathcal{L} = \lambda_{1} \mathcal{L}_{\ell_{1}} + \lambda_{2} \mathcal{L}_{adv} + \lambda_{3} \mathcal{L}_{perc} + \lambda_{4} \mathcal{L}_{style}
\end{equation}
where $\lambda_{1}$, $\lambda_{2}$, $\lambda_{3}$, and $\lambda_{4}$ are the tradeoff parameters.
In our implementation, we empirically set $\lambda_{1} = 1$, $\lambda_{2} = 0.1$, $\lambda_{3} = 0.05$ and $\lambda_{4} = 120$.

\vspace{-0.4em}
\section{Experiments}
\vspace{-0.4em}
Experiments are conducted for evaluating our LBAM on two datasets, \ie, Paris StreetView \cite{doersch2015makes} and Places (Places365-standard) \cite{zhou2017places}, which have been extensively adopted in image inpainting literature~\cite{pathakCVPR16context,Yan_2018_Shift,Yang_2017_CVPR,yu2018generative}.
For Paris StreetView, we use its original splits, $14,900$ images for training, and $100$ images for testing.
In our experiments, $100$ images are randomly selected and removed from the training set to form our validation set.
As for Places, we randomly select $10$ categories from the $365$ categories, and use all the $5,000$ images per category from the original training set to form our training set of $50,000$ images.
Moreover, we divide the original validation set from each category of $1,000$ images into two equal non-overlapped sets of $500$ images respectively for validation and testing.
Our LBAM takes $\sim70$ ms for processing a $256\times256$ image, $5\times$ faster than Context Attention~\cite{yu2018generative} ($\sim400ms$) and $\sim3\times$ faster than Global\&Local(GL)~\cite{IizukaGL} ($\sim200ms$).

In our experiments, all the images are resized where the minimal height or width is $350$, and then randomly cropped to the size of $256 \times 256$.
Data augmentation such as flipping is adopted during training.
We generate $18,000$ masks with random shape, and $12,000$ masks from~\cite{partialconv2017} for training and testing.
Our model is optimized using the ADAM algorithm~\cite{AdamOptim} with initial learning rate of $1e-4$ and $\beta=0.5$.
The training procedure ends after $500$ epochs, and the mini-batch size is $48$.
All the experiments are conducted on a PC equipped with $4$ parallel NVIDIA GTX 1080Ti GPUs.

\begin{figure*}[hbt]
\small
	\setlength{\tabcolsep}{2.0pt}
	\centering
	\vspace{-1.2em}
	\begin{tabular}{cccccccc}
		
		\includegraphics[width=.13\textwidth]{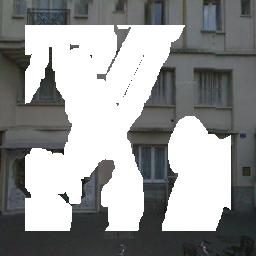}  &
		\includegraphics[width=.13\textwidth]{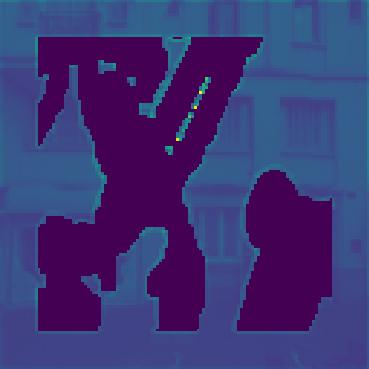}  &
		\includegraphics[width=.13\textwidth]{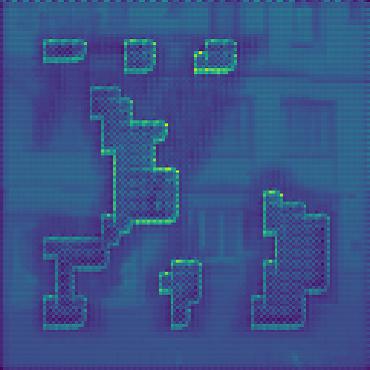}  &
		\includegraphics[width=.13\textwidth]{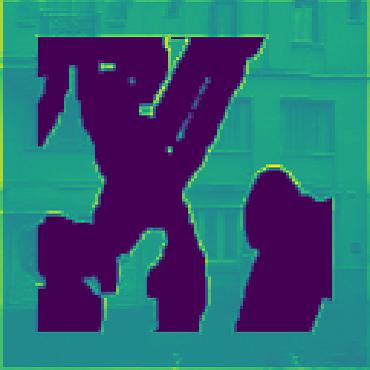}  &
		\includegraphics[width=.13\textwidth]{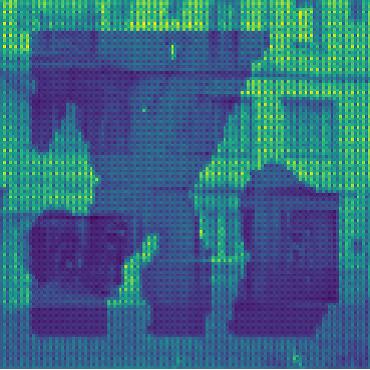}  &		
		\includegraphics[width=.13\textwidth]{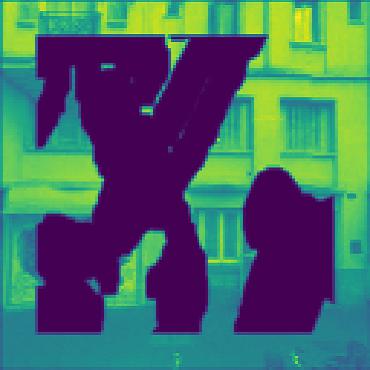}  &
		\includegraphics[width=.13\textwidth]{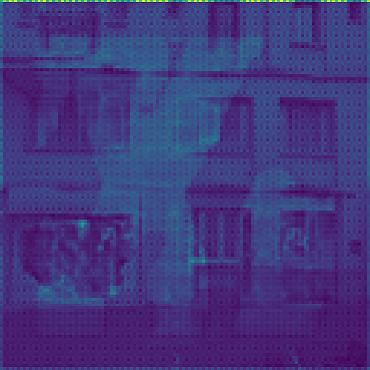}  &
		\includegraphics[width=.025\textwidth, height=.13\textwidth]{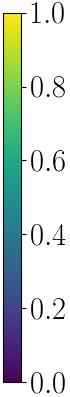}  \\
		
		(a) & (b) & (c) &(d) & (e) & (f) & (g) \\
		
	\end{tabular}
	\caption{Visualization of features from the first encoder layer and $13$-th decoder layer. (a) Input, (b)(c) Ours(unlearned), (d)(e) Ours(forward), (f)(g) Ours(full).}
	\label{fig:heatMaps}
	\vspace{-0.0em}
\end{figure*}
\begin{figure*}[hbt]
	\small
	\setlength{\tabcolsep}{2.0pt}
	\centering
	\vspace{-1.0em}
	\begin{tabular}{cccccccc}
		
		\includegraphics[width=.13\textwidth]{heatmaps/input}  &
		\includegraphics[width=.13\textwidth]{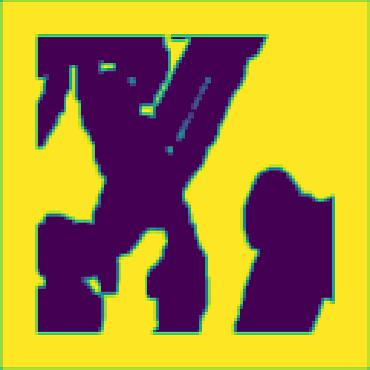}  &
		\includegraphics[width=.13\textwidth]{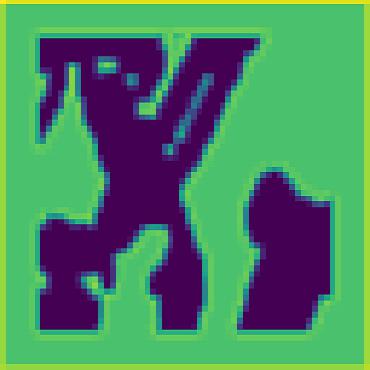}  &
		\includegraphics[width=.13\textwidth]{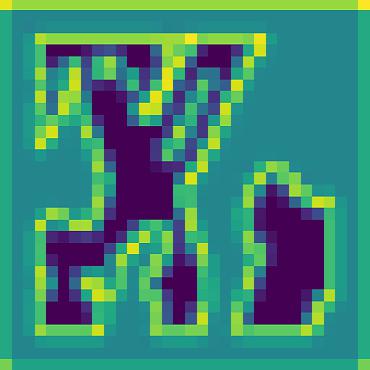}  &
		\includegraphics[width=.13\textwidth]{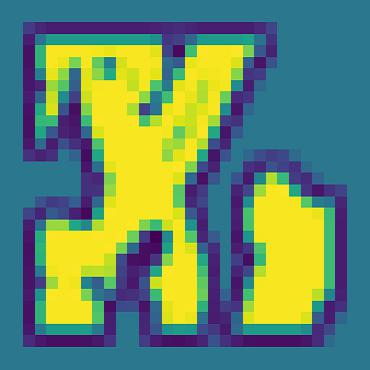}  &		
		\includegraphics[width=.13\textwidth]{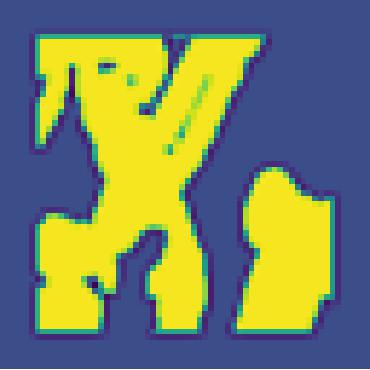}  &
		\includegraphics[width=.13\textwidth]{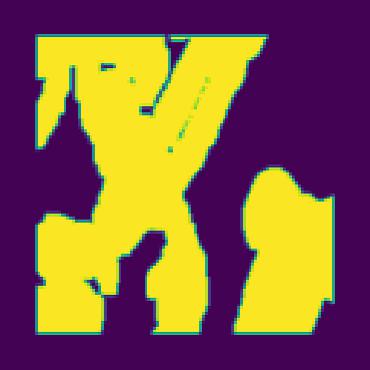}  &
		\includegraphics[width=.025\textwidth, height=.13\textwidth]{heatmaps/colorbar}  \\
		
		(a) & (b) & (c) &(d) & (e) & (f) & (g) \\
		
	\end{tabular}
	\caption{Visualization of updated masks after activation function $g_A(\cdot)$ for forward and reverse attention maps. (a) Input, (b)(c)(d) forward masks from the first three (1,2,3) layers, (e)(f)(g) reverse masks from the last three (11, 12, 13) layers.}
	\label{fig:maskMaps}
	\vspace{-1.8em}
\end{figure*}

\vspace{-0.4em}
\subsection{Comparison with State-of-the-arts}
\vspace{-0.4em}

Our LBAM is compared with four state-of-the-art methods, \ie, Global\&Local~\cite{IizukaGL}, PatchMatch~\cite{Barnes:2009:PAR}, Context Attention~\cite{yu2018generative}, and PConv~\cite{partialconv2017}.

\vspace{0.1em}
\noindent\textbf{Evaluation on Paris StreetView and Places.}
Fig.~\ref{fig:paris} and Fig.~\ref{fig:places} show the results by our LBAM and the competing methods.
Global\&Local~\cite{IizukaGL} is limited in handling irregular holes, producing many matchless and meaningless textures.
PatchMatch~\cite{Barnes:2009:PAR} performs poorly for recovering complex structures, and the results are not consistent with surrounding context.
For some complex and irregular holes, context attention~\cite{yu2018generative} still generates blurry results and may produce unwanted artifacts.
PConv~\cite{partialconv2017} is effective in handling irregular holes, but over-smoothing results are still inevitable in some regions.
In contrast, our LBAM performs well generating visually more plausible results with fine-detailed, and realistic textures.

\vspace{0.1em}
\noindent\textbf{Quantitative Evaluation.}
We also compare our LBAM quantitatively with the competing methods on Places~\cite{zhou2017places} with mask ratio $(0.1, 0.2]$, $(0.2, 0.3]$, $(0.3, 0.4]$ and $(0.4, 0.5]$.
From Table~\ref{table:quantitative_places}, our LBAM performs favorably in terms PSNR, SSIM, and mean $\ell_1$ loss, especially when the mask ratio is higher than $0.3$.

\begin{table}[hbt]\small
	\vspace{-0.8em}
	\caption{Quantitative comparison on Places. Results of PConv* are taken from~\cite{partialconv2017}.} 
	\centering 
	\setlength{\tabcolsep}{0.99mm}{
	\begin{tabular}{l|c|c|c|c|c|c} %
		\hline
		& Mask & GL~\cite{IizukaGL} & PM~\cite{Barnes:2009:PAR} & CA~\cite{yu2018generative} & PConv*~\cite{partialconv2017} & Ours \\
		\hline
		\multirow{4}*{\rotatebox{90}{PSNR}} & (0.1-0.2] & 23.36 & 26.67 & 26.27 & 28.32 & \textbf{28.51}\\
		\cline{2-7}
		~ & (0.2, 0.3] & 20.53 & 24.21 & 23.56 & 25.25 & \textbf{25.59} \\
		\cline{2-7}
		~ & (0.3, 0.4] & 19.37 & 21.95 & 21.20 & 22.89 & \textbf{23.31} \\
		\cline{2-7}
		~ & (0.4, 0.5] & 17.86 & 20.02 & 19.95 & 21.38 & \textbf{21.66} \\
		\hline
		\multirow{4}*{\rotatebox{90}{SSIM}} & (0.1-0.2] & 0.828 & 0.876 & \textbf{0.881} & 0.870 & 0.872 \\
		\cline{2-7}
		~ & (0.2, 0.3] & 0.744 & 0.763 & 0.769 & 0.779 & \textbf{0.785} \\
		\cline{2-7}
		~ & (0.3, 0.4] & 0.643 & 0.657 & 0.667 & 0.689 & \textbf{0.708} \\
		\cline{2-7}
		~ & (0.4, 0.5] & 0.545 & 0.572 & 0.563 & 0.595 & \textbf{0.602} \\
		\hline
		\multirow{4}*{\rotatebox{90}{Mean $l_1$(\%)}} & (0.1-0.2] & 2.45 & 1.43 & 2.05 & \textbf{1.09} & 1.12 \\
		\cline{2-7}
		~ & (0.2, 0.3] & 4.01 & 2.38 & 3.74 & \textbf{1.88} & 1.93 \\
		\cline{2-7}
		~ & (0.3, 0.4] & 5.86 & 3.59 & 5.65 & 2.84 & \textbf{2.55}\\
		\cline{2-7}
		~ & (0.4, 0.5] & 7.92 & 5.22 & 7.43 & 3.85 & \textbf{3.67} \\
		\hline
	\end{tabular}}
	\vspace{-1.0em}
	\label{table:quantitative_places} 
\end{table}

\begin{figure}[hbt]
	\setlength{\tabcolsep}{2.0pt}
	\vspace{-0.0em}
	\centering
	\begin{tabular}{cccc}
		\includegraphics[width=.23\linewidth]{heatmaps/input}  &
		\includegraphics[width=.23\linewidth]{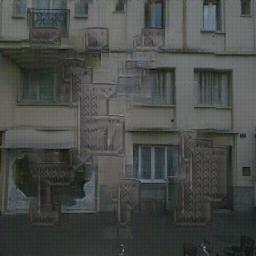}  &
		\includegraphics[width=.23\linewidth]{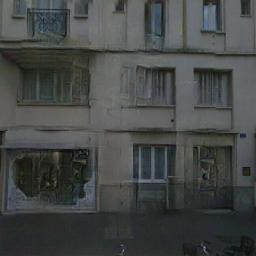}  &
		\includegraphics[width=.23\linewidth]{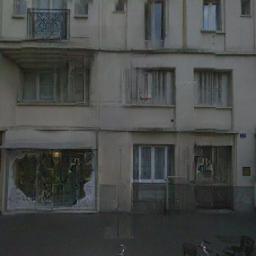}  \\
		
		{\scriptsize{(a) Input}} & {\scriptsize{(b) Ours(unlearned)}} & {\scriptsize{(c) Ours(forward)}} & {\scriptsize{(d) Ours(full)}} \\
		
	\end{tabular}
	\caption{Visual quality comparison of the effect on the learnable bidirectional attention maps.}
	\vspace{-1.6em}
	\label{fig:ablation_study}
\end{figure}

\vspace{0.1em}
\noindent\textbf{Object Removal from Real-world Images.}
Using the model trained on Places, we further evaluate LBAM on the real world object removal task.
%
%
%
Fig.~\ref{fig:real} shows the results by our LBAM, context attention~\cite{yu2018generative} and PConv~\cite{partialconv2017}.
We mask the object area either with contour shape or with rectangular bounding box.
In contrast to the competing methods, our LBAM can produce realistic and coherent contents by both global semantics and local textures.

\vspace{0.1em}
\noindent\textbf{User Study.}
Besides, user study is conducted on Paris StreetView and Places for subjective visual quality evaluation.
We randomly select $30$ images from the test set covering with different irregular holes, and the inpainting results are generated by PatchMatch~\cite{Barnes:2009:PAR}, Global\&Local~\cite{IizukaGL}, Context Attention~\cite{yu2018generative}, PConv~\cite{partialconv2017} and ours.
We invited $33$ volunteers to vote for the most visually plausible inpainting result, which is assessed by the criteria including coherency with the surrounding context, semantic structure and fine details.
For each test image, the $5$ inpainting results are randomly arranged and presented to user along with the input image.
Our LBAM has $63.2\%$ chance to win out as the most favorable result, largely surpassing PConv~\cite{partialconv2017} ($15.2\%$), PatchMatch~\cite{Barnes:2009:PAR} ($11.1\%$), Context Attention~\cite{yu2018generative} ($6.33\%$) and Global\&Local~\cite{IizukaGL} ($4.17\%$).

\vspace{-0.4em}
\subsection{Ablation Studies}
\vspace{-0.4em}
Ablation studies are conducted to compare the performance of several LBAM variants on Paris StreetView, \ie,
(i) Ours(full): the full LBAM model,
(ii) Ours(unlearned): the LBAM model where all the elements in mask convolution filters are set as $\frac{1}{16}$ because the filter size is $4\times4$, and we adopt the activation functions defined in Eqn.~(\ref{Activation_A}) and Eqn.~(\ref{Activation_M}),
(iii) Ours(forward): the LBAM model without reverse attention map,
(iv) Ours(w/o $\mathcal{L}_{adv}$): the LBAM model without (w/o) adversarial loss,
(v) Ours(Sigmoid/LReLU/ReLU/$3\times3$): the LBAM model using Sigmoid/LeakyReLU/ReLU as activation functions or $3\times3$ filter for mask updating.

Fig.~\ref{fig:heatMaps} shows the visualization of features from the first encoder layer and $13$-th decoder layer by Ours(unlearned), Ours(forward), and Ours(full).
For Ours(unlearned), blurriness and artifacts can be observed from Fig.~\ref{fig:ablation_study}(b).
%
%
Ours(forward) is beneficial to reduce the artifacts and noise, but the decoder hallucinates both holes and known regions and produces some blurry effects (see Fig.~\ref{fig:ablation_study}(c)).
%
%
In contrast, Ours(full) is effective in generating semantic structure and detailed textures (see Fig.~\ref{fig:ablation_study}(d)), and the decoder focus mainly on hallucinating holes (see Fig.~\ref{fig:heatMaps}(g)).
%
%
Table~\ref{table:quantitative_paris} gives the quantitative results of the LBAM variants on Paris StreetView, and the performance gain of Ours(full) can be explained by (1) learnable attention maps, (2) reverse attention maps, and (3) proper activation functions.


\begin{table}[hbt]\scriptsize
	\vspace{-0.8em}
	\caption{Ablation studies (PSNR/SSIM) on Paris StreetView.} 
	\centering 
	\begin{tabular}{l|c|c|c|c} %
		\hline %
		Method & (0.1, 0.2] & (0.2, 0.3] & (0.3, 0.4]  & (0.4, 0.5] \\
		\hline 
		Ours(unlearned) & 26.95/0.853 & 24.39/0.763 & 22.54/0.677 & 21.20/0.583 \\
		Ours(forward) & 27.80/0.869	& 25.13/0.775 &	23.04/0.688	& 21.76/0.598  \\
		Ours(Sigmoid) & 26.93/0.857 & 24.15/0.768 & 22.24/0.683 & 20.32/0.582  \\
		Ours(LReLU) & 26.61/0.852 & 23.59/0.762 & 20.63/0.667 & 18.38/0.562  \\
		Ours(ReLU) & 27.62/0.864 & 25.16/0.776 & 22.96/0.685 & 21.48/0.596 \\
		Ours(3x3) & \textcolor{blue}{28.74}/0.886 & 26.10/0.793 & 24.03/0.703 & 22.43/0.617 \\
		Ours(w/o $\mathcal{L}_{adv}$) & \textcolor{red}{29.19}/\textcolor{red}{0.903} & \textcolor{red}{26.55}/\textcolor{red}{0.817} & \textcolor{red}{24.46}/\textcolor{red}{0.729} & \textcolor{red}{22.70}/\textcolor{red}{0.626}  \\
		\textbf{Ours(full)} & 28.73/\textcolor{blue}{0.889} & \textcolor{blue}{26.16}/\textcolor{blue}{0.795} & \textcolor{blue}{24.26}/\textcolor{blue}{0.716}  & \textcolor{blue}{22.62}/\textcolor{blue}{0.621}  \\
		\hline 
	\end{tabular}
	\vspace{-1.5em}
	\label{table:quantitative_paris} 
\end{table}

\vspace{0.1em}
\noindent\textbf{Mask Updating.}\ Fig.~\ref{fig:maskMaps} shows the visualization of updated masks from different layers.
From the first to third layers, the masks of encoder are gradually updated to reduce the size of holes.
Analogously, from the 13-th to 11-th layers, the masks of decoder are gradually updated to reduce the size of known region.
%

%
%
%

\vspace{0.1em}
\noindent \textbf{Effect of Adversarial Loss.}
Table~\ref{table:quantitative_paris} also gives the quantitative result w/o $\mathcal{L}_{adv}$.
Albeit Ours(w/o $\mathcal{L}_{adv}$) improves PSNR and SSIM, the use of $\mathcal{L}_{adv}$ generally benefits the visual quality of the inpainting results.
The qualitative results are given in the suppl.


\vspace{-0.8em}
\section{Conclusion}
\vspace{-0.4em}
\noindent This paper proposed a learnable bidirectional attention maps (LBAM) for image inpainting.
With the introduction of learnable attention maps, our LBAM is effective in adapting to irregular holes and propagation of convolution layers.
%
%
Furthermore, reverse attention maps are presented to allow the decoder of U-Net concentrate only on filling in holes.
Experiments shows that our LBAM performs favorably against state-of-the-arts in generating sharper, more coherent and fine-detailed results.

\vspace{-0.8em}
\section*{Acknowledgement}
\vspace{-0.4em}
\noindent This work was supported in part by the NSFC grant under No. 61671182 and 61872116, and National Key Research and Development Project 2018YFC0832105.

\clearpage

{\small
\bibliographystyle{ieee_fullname}
\bibliography{egbib}
}

\clearpage

\begin{appendix}

\section*{Supplementary Material}

\subsection*{Visual comparison of several LBAM variants on Paris StreetView dataset}
We implement our bidirectional attention maps by employing an asymmetric Gaussian shaped form (Eqn.~\ref{Activation_gA}) for activation the attention map and the modified activation function (Eqn.~\ref{Activation_gM}) for updating the mask. In this material, we give visual comparison of several variants of our LBAM model, \ie, 
(i) Ours(full): the full LBAM model, 
(ii) Ours(unlearned): the LBAM model where all the elements in mask convolution filters are set as $\frac{1}{16}$ because the filter size is $4\times4$, and we adopt the activation functions defined in Eqn.~\ref{Activation_A} and Eqn.~\ref{Activation_M}, 
(iii) Ours(forward): the LBAM model without reverse attention map,
(iv) Ours(w/o $\mathcal{L}_{adv}$): the LBAM model without (w/o) adversarial loss,
(v) Ours(Sigmoid/LReLU/ReLU/$3\times3$): the LBAM model using Sigmoid/LeakyReLU/ReLU as activation functions or $3\times3$ filter for mask updating.

\begin{figure*}[hbt]
	\setlength{\tabcolsep}{2.0pt}
	\centering
	\begin{tabular}{ccccc}
		\includegraphics[width=.19\textwidth]{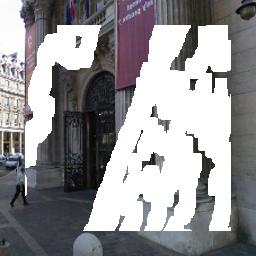}  &
		\includegraphics[width=.19\textwidth]{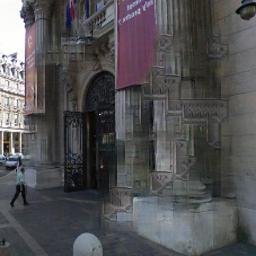}  &
		\includegraphics[width=.19\textwidth]{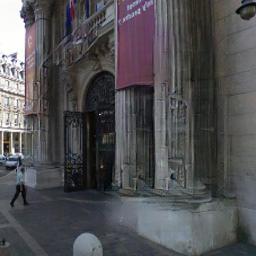}  &
		\includegraphics[width=.19\textwidth]{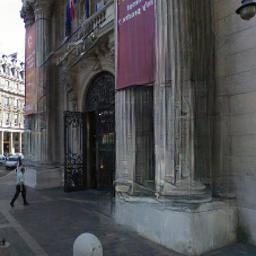}  &
		\includegraphics[width=.19\textwidth]{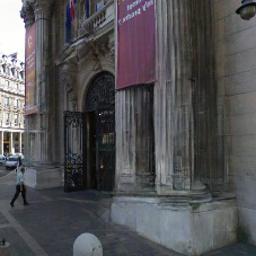}  \\
		
		\includegraphics[width=.19\textwidth]{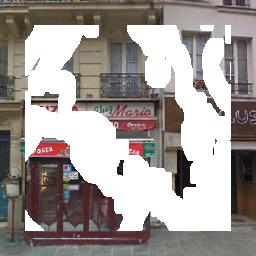}  &
		\includegraphics[width=.19\textwidth]{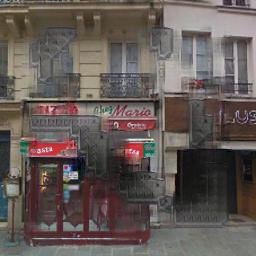}  &
		\includegraphics[width=.19\textwidth]{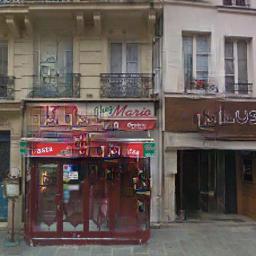}  &
		\includegraphics[width=.19\textwidth]{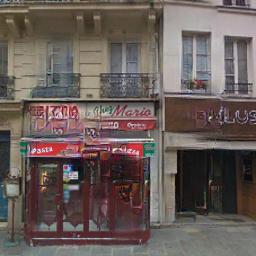}  &
		\includegraphics[width=.19\textwidth]{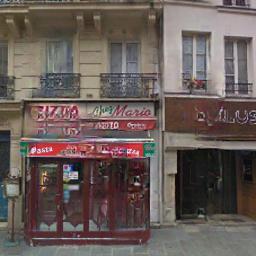}  \\
		
		\includegraphics[width=.19\textwidth]{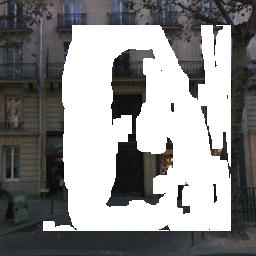}  &
		\includegraphics[width=.19\textwidth]{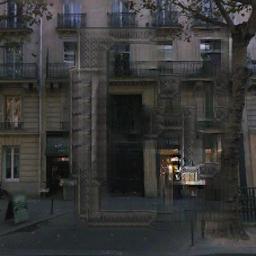}  &
		\includegraphics[width=.19\textwidth]{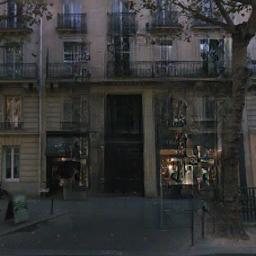}  &
		\includegraphics[width=.19\textwidth]{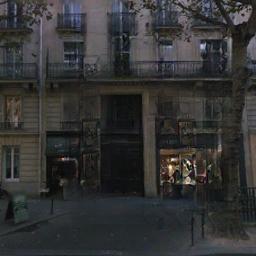}  &
		\includegraphics[width=.19\textwidth]{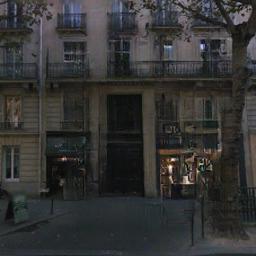}  \\
		
		Input & (a) Ours(unlearned) & (b) Ours(forward) & (c) Ours(w/o $\mathcal{L}_{adv}$) & (d)  Ours(full) \\
		
	\end{tabular}
	\caption{Visual comparison of variants (i) to (iii) of our LBAM model. From left to right are: Input, (a) Ours with unlearned model, (b) Ours without reverse attention map, (c) Our without (w/o) adversarial loss, (d) our full LBAM model. All images are scaled to $256 \times 256$.}
	\label{fig:study1}
	\vspace{-0.8em}
\end{figure*}

Fig.~\ref{fig:study1} shows qualitatively comparison over variants (i) to (iv). Ours (forward) model benefits from learnable attention map and helps reduce reduce the artifacts and noise of unlearned one, see Fig.~\ref{fig:study1}(a) and (b). But its decoder hallucinates both holes and known regions and produces some blurry effects compared to our full model with learnable reverse attention map Fig.~\ref{fig:study1}(d). 

The qualitative comparison in ablation studies with the effect of GAN loss is shown in Fig.~\ref{fig:study1}(c) and (d). The inpainted results of our LBAM model without adversarial loss (Fig.~\ref{fig:study1}(c)), are much better than the unlearned model Fig.~\ref{fig:study1}(a), and somehow clearer in producing details than ours without reverse attention map which applied GAN loss. Our LBAM full model (Fig.~\ref{fig:study1}(d)) benefits from GAN loss, is superior in giving fine-detailed structures and capturing global semantics. 

\begin{figure*}[hbt]
	\setlength{\tabcolsep}{2.0pt}
	\centering
	\begin{tabular}{cccccc}
		\includegraphics[width=.16\textwidth]{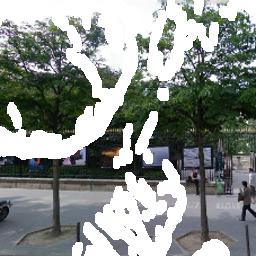}  &
		\includegraphics[width=.16\textwidth]{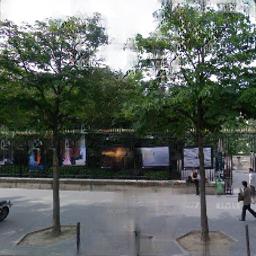}  &
		\includegraphics[width=.16\textwidth]{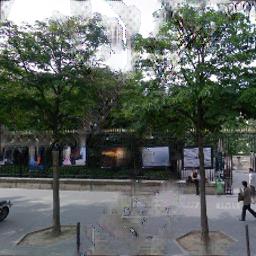}  &
		\includegraphics[width=.16\textwidth]{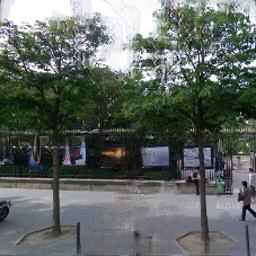}  &
		\includegraphics[width=.16\textwidth]{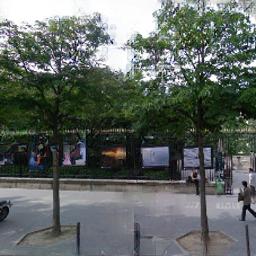}  &
		\includegraphics[width=.16\textwidth]{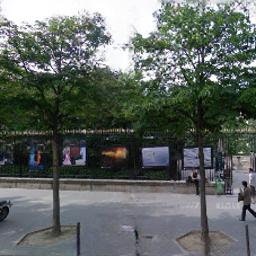}  \\
		
		\includegraphics[width=.16\textwidth]{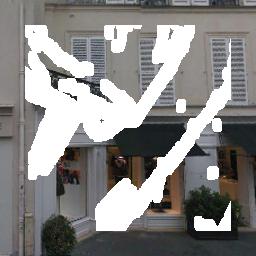}  &
		\includegraphics[width=.16\textwidth]{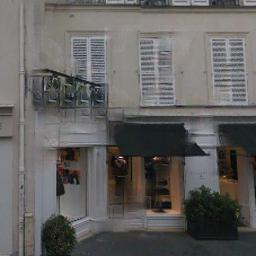}  &
		\includegraphics[width=.16\textwidth]{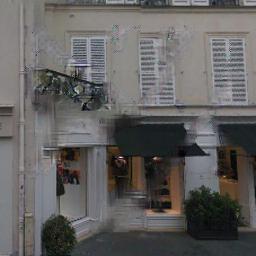}  &
		\includegraphics[width=.16\textwidth]{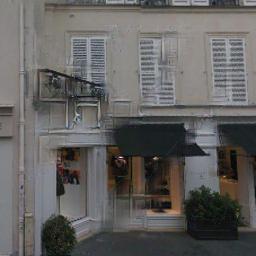}  &
		\includegraphics[width=.16\textwidth]{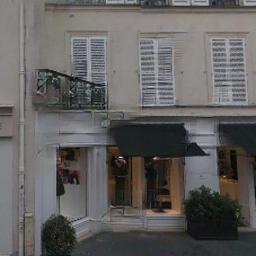}  &
		\includegraphics[width=.16\textwidth]{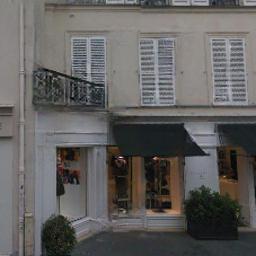}  \\
		
		Input & (a) & (b) & (c)  & (d) & (e) Ours \\
		
	\end{tabular}
	\caption{Visual comparison of different activation functions or $3\times3$ filters on the bidirectional attention maps. From left to right are: Input, (a) Sigmoid as activation function, (b) Leaky ReLU with slope of $0.2$ as activation function, (c) ReLU, (e) $3\times3$ filter for mask updating, and (e) our full LBAM model. All images are scaled to $256 \times 256$.}
	\label{fig:study2}
	\vspace{-0.8em}
\end{figure*}

The visual comparison of different activation functions or $3\times3$ filter for mask updating are shown in Fig.~\ref{fig:study2}.

\begin{figure*}[hbt]
	\setlength{\tabcolsep}{2.0pt}
	\centering
	\begin{tabular}{cccccc}
		\includegraphics[width=.16\textwidth]{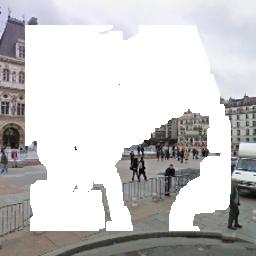}  &
		\includegraphics[width=.16\textwidth]{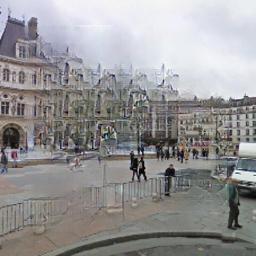}  &
		\includegraphics[width=.16\textwidth]{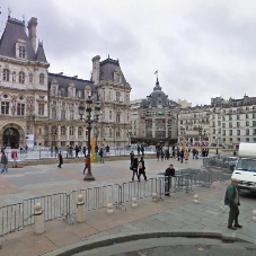}  &
		
				\includegraphics[width=.16\textwidth]{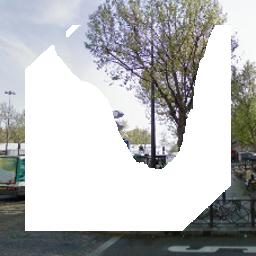}  &
				\includegraphics[width=.16\textwidth]{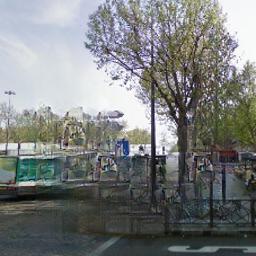}  &
				\includegraphics[width=.16\textwidth]{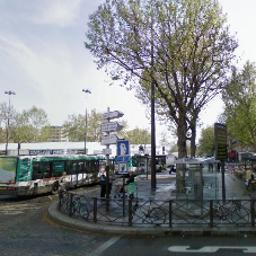}  \\	
		
				\includegraphics[width=.16\textwidth]{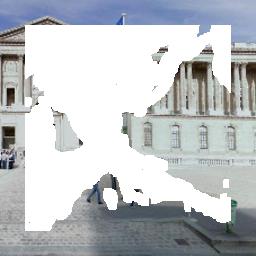}  &
				\includegraphics[width=.16\textwidth]{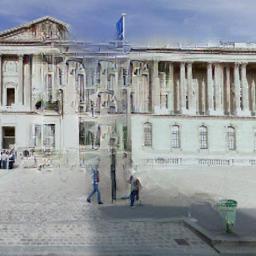}  &
				\includegraphics[width=.16\textwidth]{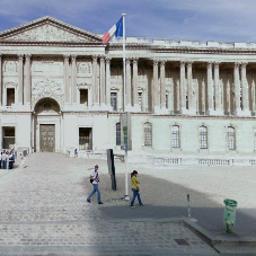}  &
		\includegraphics[width=.16\textwidth]{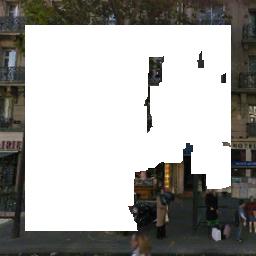}  &
		\includegraphics[width=.16\textwidth]{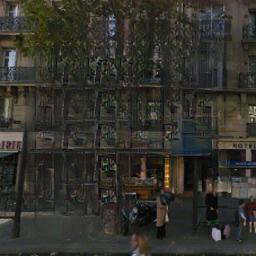}  &
		\includegraphics[width=.16\textwidth]{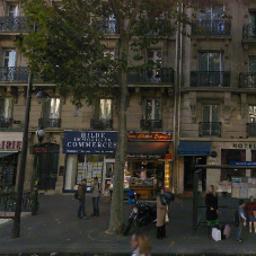}  \\
		
		\includegraphics[width=.16\textwidth]{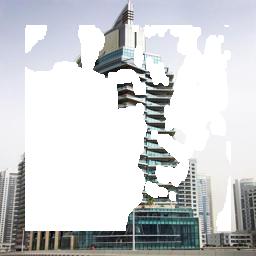}  &
		\includegraphics[width=.16\textwidth]{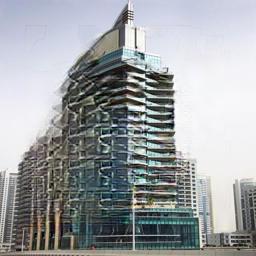}  &
		\includegraphics[width=.16\textwidth]{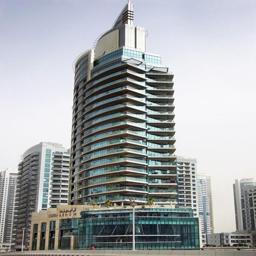}  &
		\includegraphics[width=.16\textwidth]{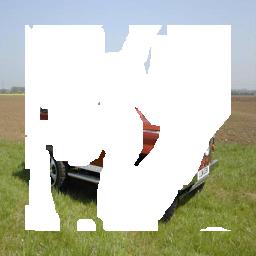}  &
		\includegraphics[width=.16\textwidth]{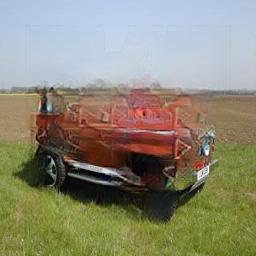}  &
		\includegraphics[width=.16\textwidth]{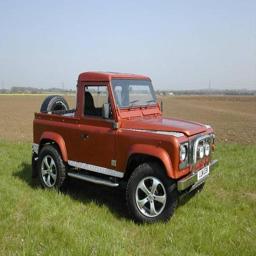}  \\
		Input & Ours & Ground Truth &  Input & Ours & Ground Truth \\
		
	\end{tabular}
	\caption{Failure cases of our LBAM model. Each group is ordered as input image, our result and ground truth. All images are scaled to $256 \times 256$.}
	\label{fig:failure_cases}
	\vspace{-1.8em}
\end{figure*}

\noindent\textbf{Failure cases.} Fig.~\ref{fig:failure_cases} shows some failure cases of our LBAM model. Our model struggles to recover the high-frequency details while the damaged areas are too large or the background objects are too complex. In some cases, the mask covers a large portion of a specific object, like a car, it is still difficult for our LBAM model to recover the original shape.

\subsection*{Model Architectures}

\subsubsection*{Architecture of Our Learnable Bidirectional Attention Map}
The learnable bidirectional attention model takes the damaged image, the mask $M^{in}$ and the reverse mask $1 - M^{in}$ as input. We adopt the basic U-Net structure with $14$ layers, and both encoder and decoder consists of $7$ layers. The features are normalized by the learnable bidirectional attention maps through element-wise product. We use convolution filters of size $4 \times 4$, $\mbox{stride}=2$, $\mbox{padding}=1$ for all layers including the bidirectional attention maps. 

The forward attention map takes the mask $M^{in}$ as input, it contains $7$ layers, and the reverse attention map takes the reverse mask $1 - M^{in}$ as input, which consists of $6$ layers. We adopt an asymmetric Gaussian-shaped form as activation function ($g_{A}(\cdot)$ of Eqn.~\ref{Activation_gA}) for activating the attention map and a modified ReLU based activating function ($g_{M}(\cdot)$ of Eqn.~\ref{Activation_gM}) for updating mask maps. 
In consideration of the skip connection of the U-Net structure, the symmetric forward and reverse attention maps are concatenated for normalizing the connected features of the corresponding layer in the decoder, under Eqn.~\ref{PConv-2-r}.
Besides, batch normalization and Leaky ReLU non-linearity are used to the features after attention re-normalization.
The last layer of our LBAM model are directly de-convoluted with filters of size $4 \times 4$, $\mbox{stride}=2$, $\mbox{padding}=1$, followed by a tanh non-linear activation. More details about our model is given in Table~\ref{table:generator}. Note that each activation function $g_{A}(\cdot)$ and mask updating term $g_{M}(\cdot)$ are unique for each layer, and they do not share parameters among layers. 

\vspace{-1.0em}
\subsubsection*{Architecture of the Discriminator}
\vspace{-0.5em}
The discriminator is trained to produce adversarial loss for minimizing the distance between the generated images and the real data distributions. In our work, we use a two-column discriminator with one column takes the remained area of inpainted result or a ground-truth image, and another column takes the missing holes of inpainted result or a ground-truth image as input. The two-column discriminator consists of $7$ layers, the two parallel features are emerged after $6_{th}$ layer at the resolution of $4\times4$. We specifically use convolution layer with filters size of $4\times4$, $\mbox{stride}=2$ and $\mbox{padding}=1$, except the last layer with $\mbox{stride}=0$. We use sigmoid non-linear activation function at last layer, while the leaky ReLU with slope of 0.2 for other layers. Table~\ref{table:discriminator} provides a more details of the discriminator.

\subsection*{More Comparisons on Paris StreetView and Places }
\begin{figure*}[hbt]
	\setlength{\tabcolsep}{2.0pt}
	\centering
	\begin{tabular}{cccccc}
		\includegraphics[width=.16\textwidth]{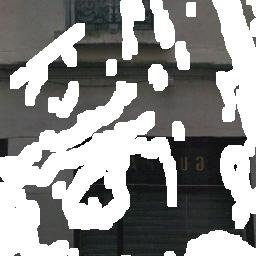}  &
		\includegraphics[width=.16\textwidth]{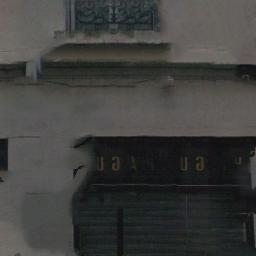}  &
		\includegraphics[width=.16\textwidth]{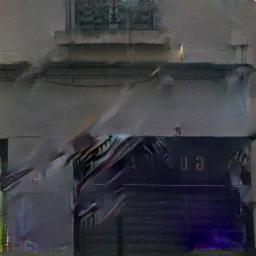}  &
		\includegraphics[width=.16\textwidth]{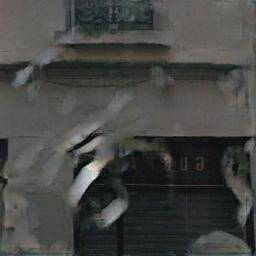}  &
		\includegraphics[width=.16\textwidth]{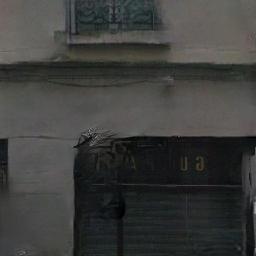}  &
		\includegraphics[width=.16\textwidth]{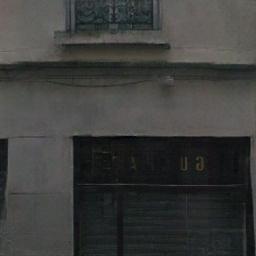}  \\
		
		\includegraphics[width=.16\textwidth]{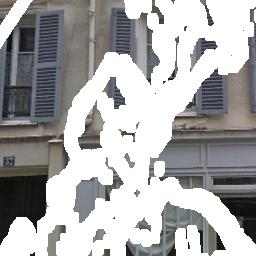}  &
		\includegraphics[width=.16\textwidth]{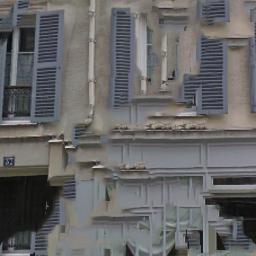}  &
		\includegraphics[width=.16\textwidth]{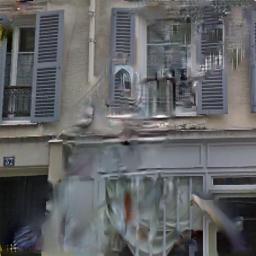}  &
		\includegraphics[width=.16\textwidth]{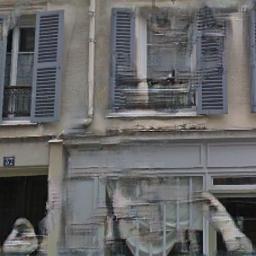}  &
		\includegraphics[width=.16\textwidth]{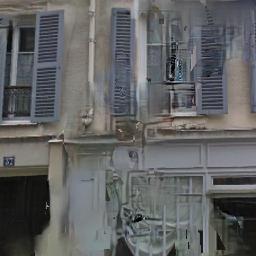}  &
		\includegraphics[width=.16\textwidth]{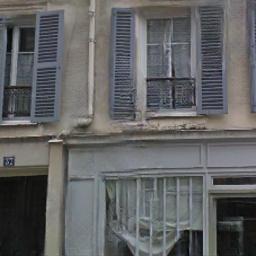}  \\

		\includegraphics[width=.16\textwidth]{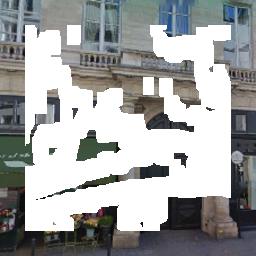}  &
		\includegraphics[width=.16\textwidth]{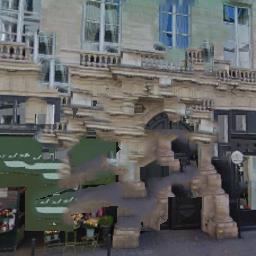}  &
		\includegraphics[width=.16\textwidth]{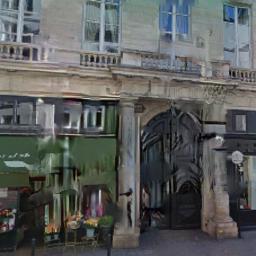}  &
		\includegraphics[width=.16\textwidth]{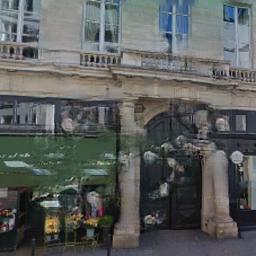}  &
		\includegraphics[width=.16\textwidth]{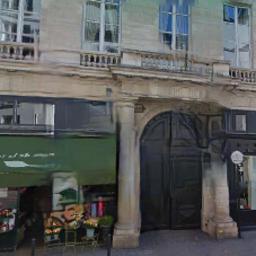}  &
		\includegraphics[width=.16\textwidth]{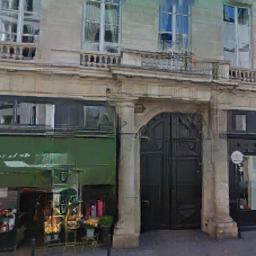}  \\

		\includegraphics[width=.16\textwidth]{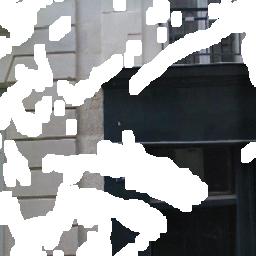}  &
		\includegraphics[width=.16\textwidth]{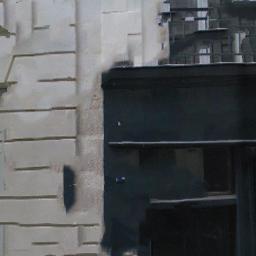}  &
		\includegraphics[width=.16\textwidth]{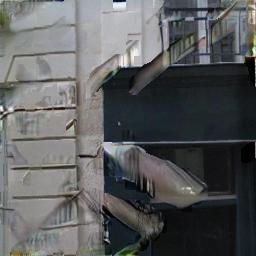}  &
		\includegraphics[width=.16\textwidth]{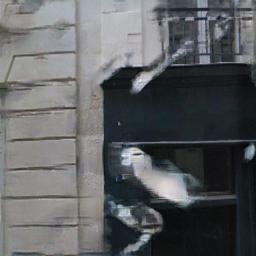}  &
		\includegraphics[width=.16\textwidth]{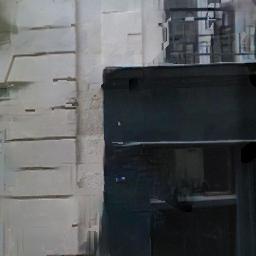}  &
		\includegraphics[width=.16\textwidth]{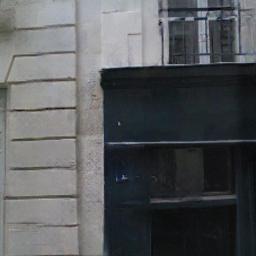}  \\
		
		\includegraphics[width=.16\textwidth]{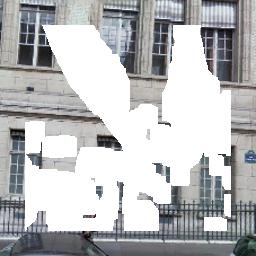}  &
		\includegraphics[width=.16\textwidth]{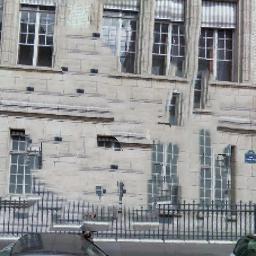}  &
		\includegraphics[width=.16\textwidth]{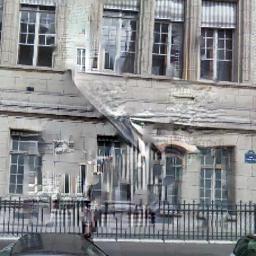}  &
		\includegraphics[width=.16\textwidth]{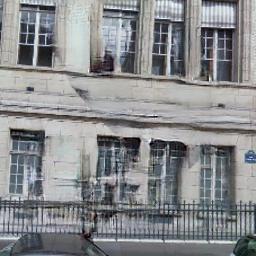}  &
		\includegraphics[width=.16\textwidth]{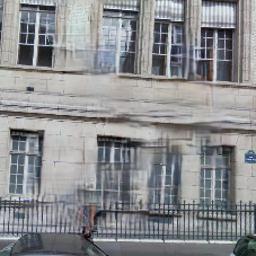}  &
		\includegraphics[width=.16\textwidth]{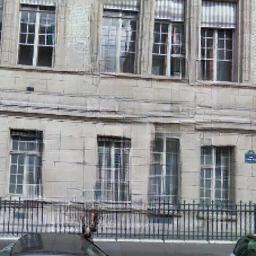}  \\
		
		\includegraphics[width=.16\textwidth]{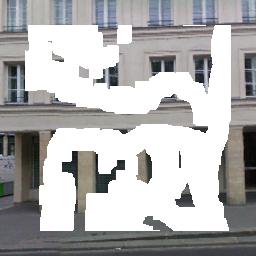}  &
		\includegraphics[width=.16\textwidth]{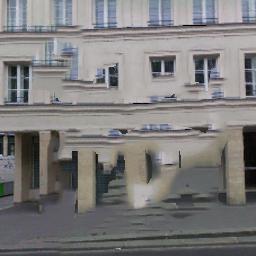}  &
		\includegraphics[width=.16\textwidth]{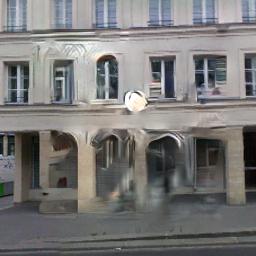}  &
		\includegraphics[width=.16\textwidth]{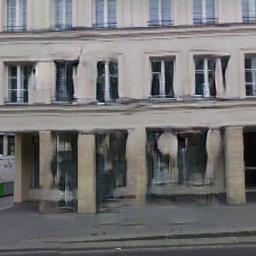}  &
		\includegraphics[width=.16\textwidth]{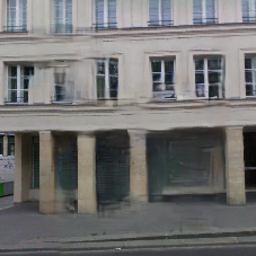}  &
		\includegraphics[width=.16\textwidth]{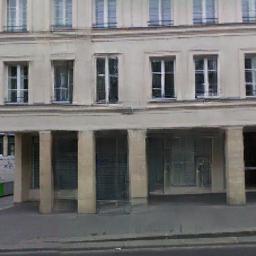}  \\
		
		\includegraphics[width=.16\textwidth]{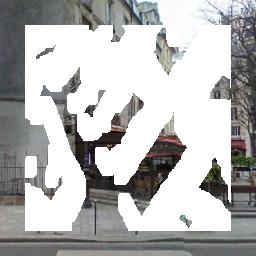}  &
		\includegraphics[width=.16\textwidth]{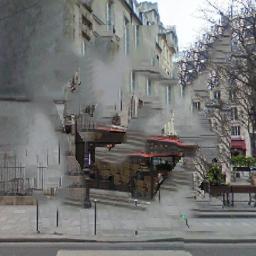}  &
		\includegraphics[width=.16\textwidth]{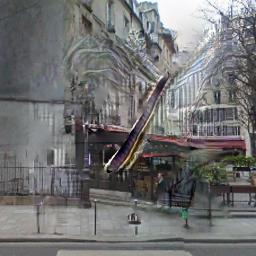}  &
		\includegraphics[width=.16\textwidth]{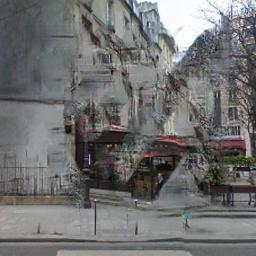}  &
		\includegraphics[width=.16\textwidth]{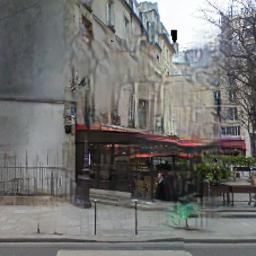}  &
		\includegraphics[width=.16\textwidth]{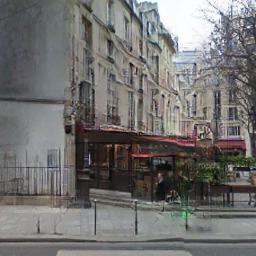}  \\

		Input & PM~\cite{Barnes:2009:PAR} & GL~\cite{IizukaGL} & CA~\cite{yu2018generative} & PConv~\cite{partialconv2017} & Ours \\
		
	\end{tabular}
	\caption{Qualitative comparison on Paris StreetView dataset. Comparison with PatchMatch (PM)~\cite{Barnes:2009:PAR}, Global\&Local (GL)~\cite{IizukaGL}, Context Attention (CA)~\cite{yu2018generative}, and Partial Convolution (PConv)~\cite{partialconv2017}. All images are scaled to $256 \times 256$.}
	\label{fig:paris1}
\end{figure*}

\begin{figure*}[hbt]
	\setlength{\tabcolsep}{2.0pt}
	\centering
	\begin{tabular}{cccccc}
		
		\includegraphics[width=.16\textwidth]{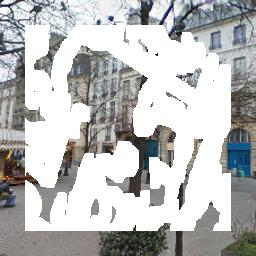}  &
		\includegraphics[width=.16\textwidth]{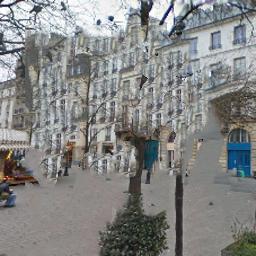}  &
		\includegraphics[width=.16\textwidth]{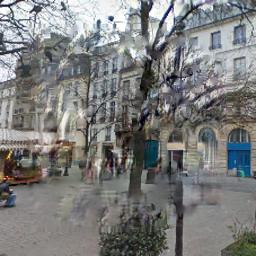}  &
		\includegraphics[width=.16\textwidth]{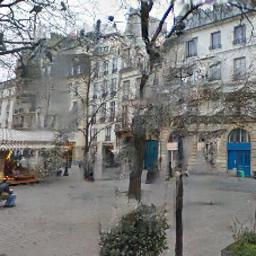}  &
		\includegraphics[width=.16\textwidth]{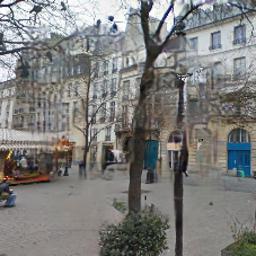}  &
		\includegraphics[width=.16\textwidth]{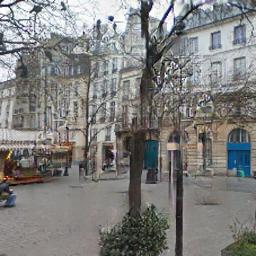}  \\
		
		\includegraphics[width=.16\textwidth]{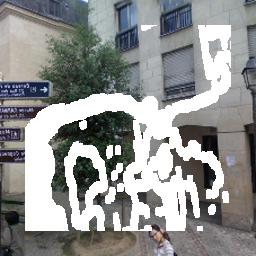}  &
		\includegraphics[width=.16\textwidth]{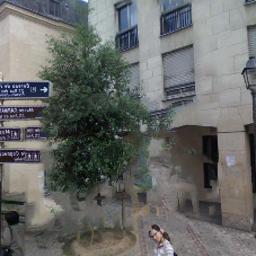}  &
		\includegraphics[width=.16\textwidth]{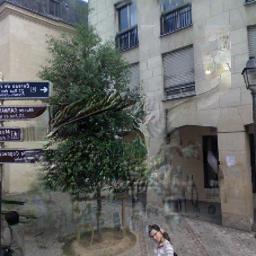}  &
		\includegraphics[width=.16\textwidth]{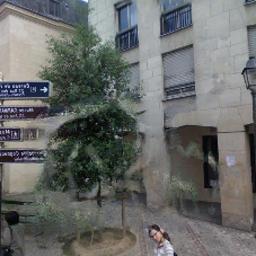}  &
		\includegraphics[width=.16\textwidth]{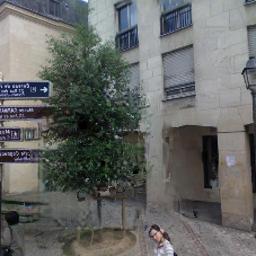}  &
		\includegraphics[width=.16\textwidth]{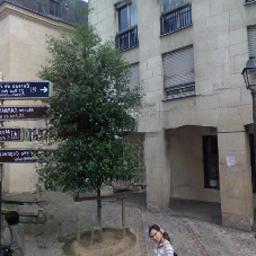}  \\

		\includegraphics[width=.16\textwidth]{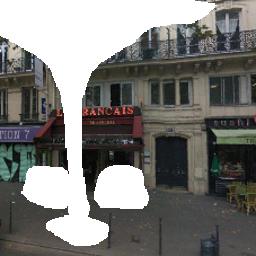}  &
		\includegraphics[width=.16\textwidth]{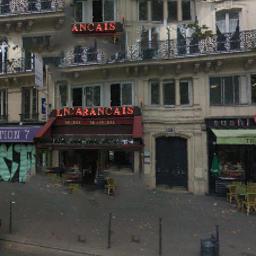}  &
		\includegraphics[width=.16\textwidth]{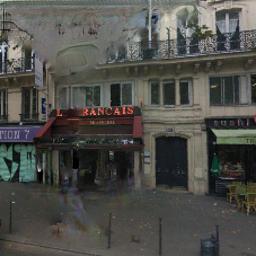}  &
		\includegraphics[width=.16\textwidth]{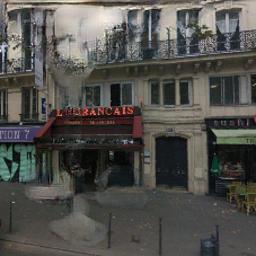}  &
		\includegraphics[width=.16\textwidth]{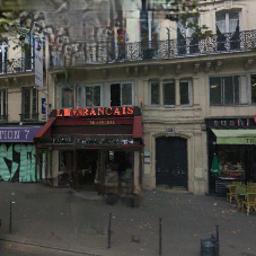}  &
		\includegraphics[width=.16\textwidth]{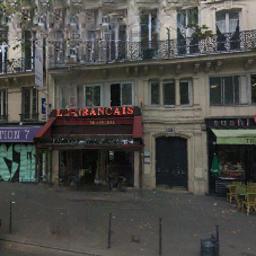}  \\
		
		
		\includegraphics[width=.16\textwidth]{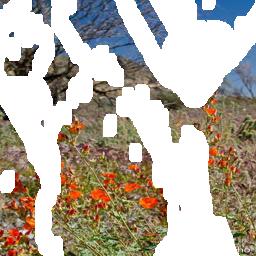}  &
		\includegraphics[width=.16\textwidth]{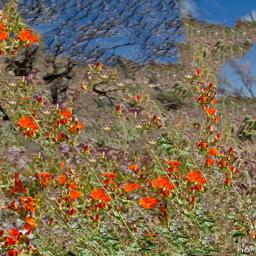}  &
		\includegraphics[width=.16\textwidth]{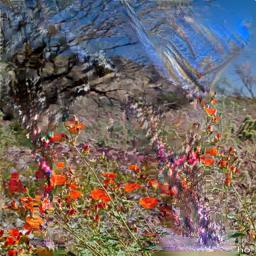}  &
		\includegraphics[width=.16\textwidth]{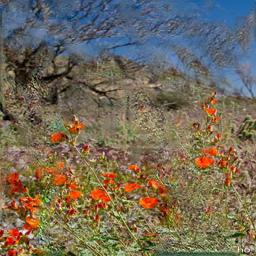}  &
		\includegraphics[width=.16\textwidth]{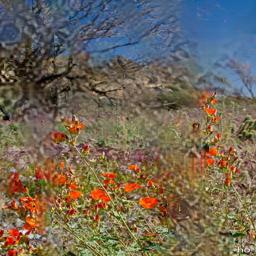}  &
		\includegraphics[width=.16\textwidth]{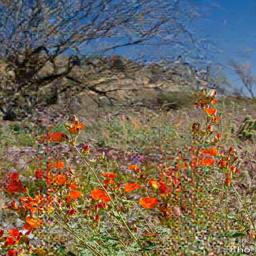}  \\	
		
		\includegraphics[width=.16\textwidth]{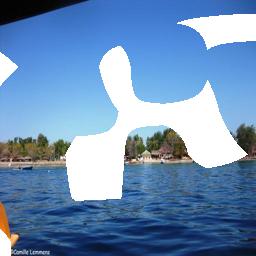}  &
		\includegraphics[width=.16\textwidth]{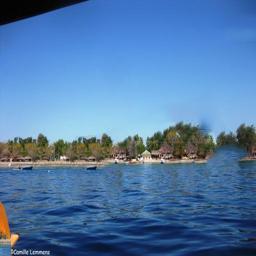}  &
		\includegraphics[width=.16\textwidth]{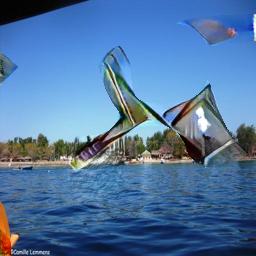}  &
		\includegraphics[width=.16\textwidth]{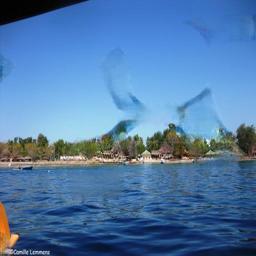}  &
		\includegraphics[width=.16\textwidth]{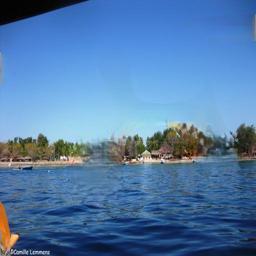}  &
		\includegraphics[width=.16\textwidth]{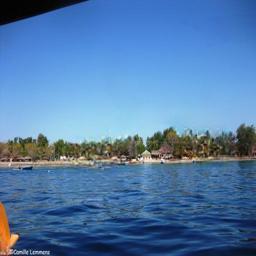}  \\
		
		\includegraphics[width=.16\textwidth]{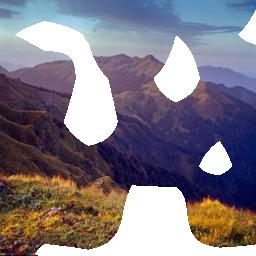}  &
		\includegraphics[width=.16\textwidth]{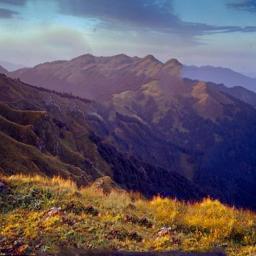}  &
		\includegraphics[width=.16\textwidth]{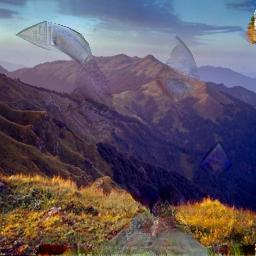}  &
		\includegraphics[width=.16\textwidth]{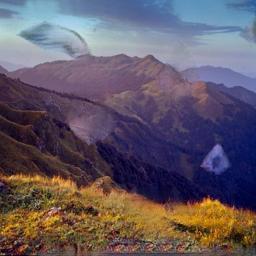}  &
		\includegraphics[width=.16\textwidth]{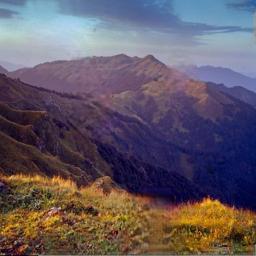}  &
		\includegraphics[width=.16\textwidth]{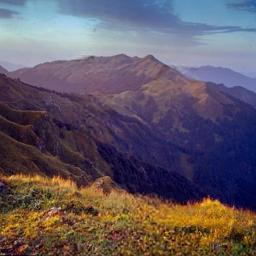}  \\
		
		\includegraphics[width=.16\textwidth]{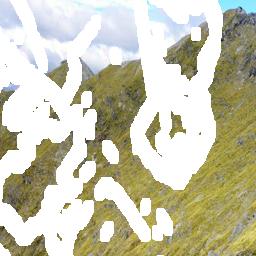}  &
		\includegraphics[width=.16\textwidth]{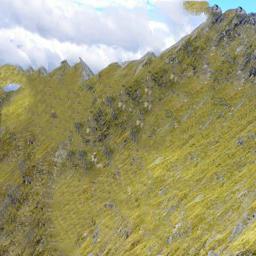}  &
		\includegraphics[width=.16\textwidth]{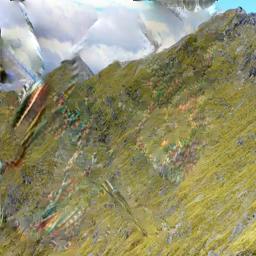}  &
		\includegraphics[width=.16\textwidth]{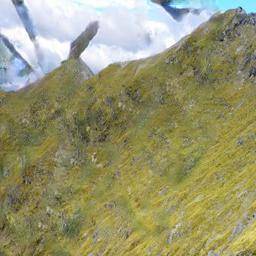}  &
		\includegraphics[width=.16\textwidth]{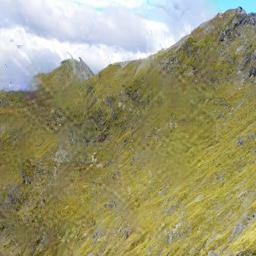}  &
		\includegraphics[width=.16\textwidth]{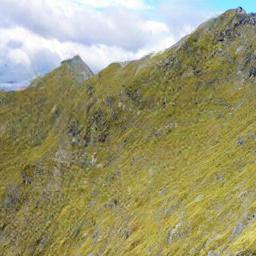}  \\

		Input & PM~\cite{Barnes:2009:PAR} & GL~\cite{IizukaGL} & CA~\cite{yu2018generative} & PConv~\cite{partialconv2017} & Ours \\
		
	\end{tabular}
	\caption{Qualitative comparison on Paris StreetView dataset. Comparison with PatchMatch (PM)~\cite{Barnes:2009:PAR}, Global\&Local (GL)GL~\cite{IizukaGL}, Context Attention (CA)~\cite{yu2018generative}, and Partial Convolution (PConv)~\cite{partialconv2017}. First three rows are from Paris StreetView dataset and the last four rows are from Places dataset. All images are scaled to $256 \times 256$.}
	\label{fig:paris2}
\end{figure*}

\begin{figure*}[hbt]
	\setlength{\tabcolsep}{2.0pt}
	\centering
	\begin{tabular}{cccccc}

		\includegraphics[width=.16\textwidth]{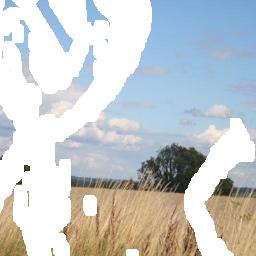}  &
		\includegraphics[width=.16\textwidth]{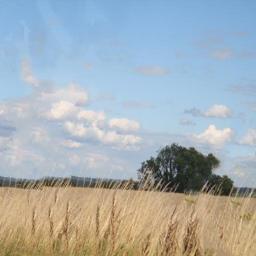}  &
		\includegraphics[width=.16\textwidth]{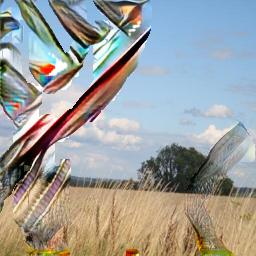}  &
		\includegraphics[width=.16\textwidth]{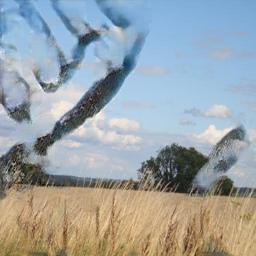}  &
		\includegraphics[width=.16\textwidth]{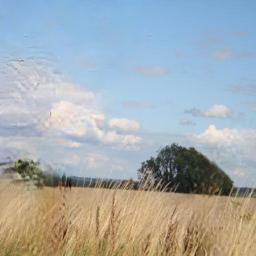}  &
		\includegraphics[width=.16\textwidth]{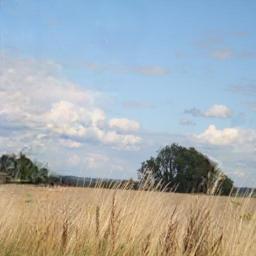}  \\
		\includegraphics[width=.16\textwidth]{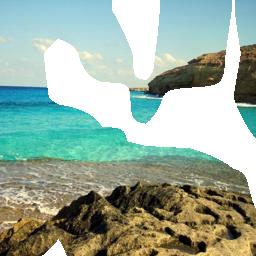}  &
		\includegraphics[width=.16\textwidth]{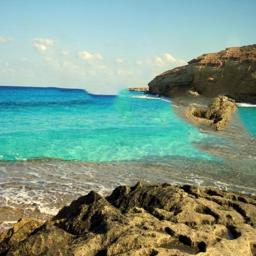}  &
		\includegraphics[width=.16\textwidth]{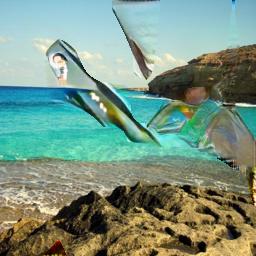}  &
		\includegraphics[width=.16\textwidth]{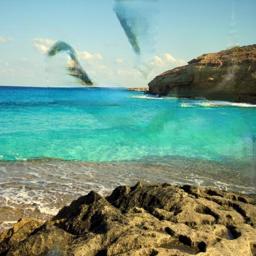}  &
		\includegraphics[width=.16\textwidth]{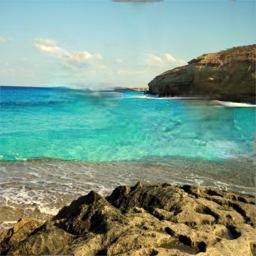}  &
		\includegraphics[width=.16\textwidth]{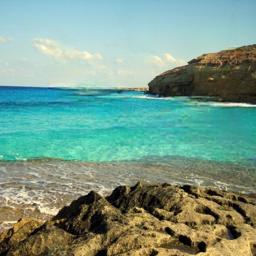}  \\
		\includegraphics[width=.16\textwidth]{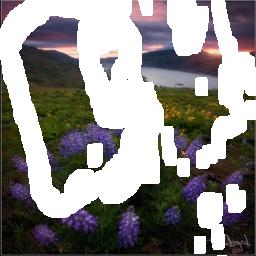}  &
		\includegraphics[width=.16\textwidth]{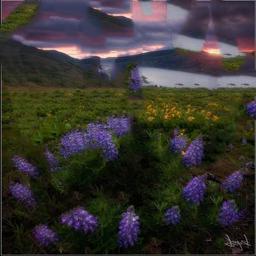}  &
		\includegraphics[width=.16\textwidth]{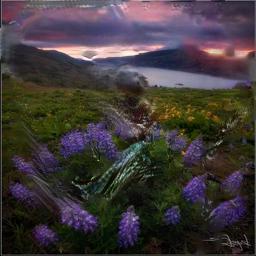}  &
		\includegraphics[width=.16\textwidth]{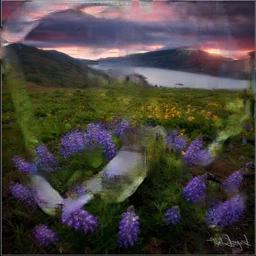}  &
		\includegraphics[width=.16\textwidth]{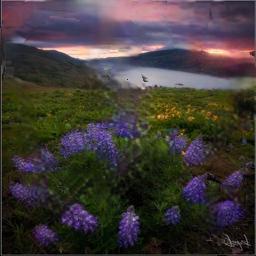}  &
		\includegraphics[width=.16\textwidth]{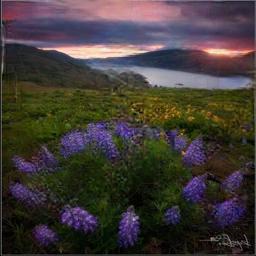}  \\
		
		\includegraphics[width=.16\textwidth]{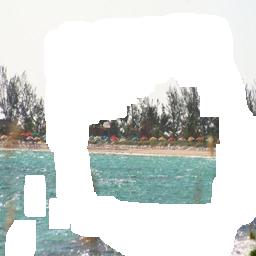}  &
		\includegraphics[width=.16\textwidth]{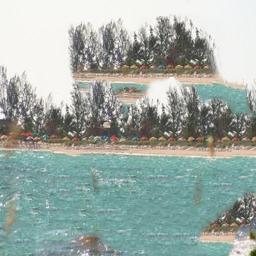}  &
		\includegraphics[width=.16\textwidth]{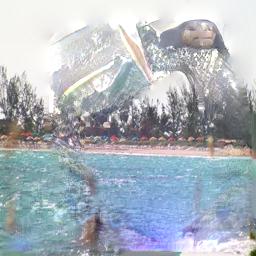}  &
		\includegraphics[width=.16\textwidth]{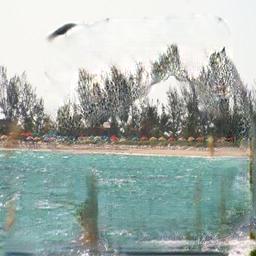}  &
		\includegraphics[width=.16\textwidth]{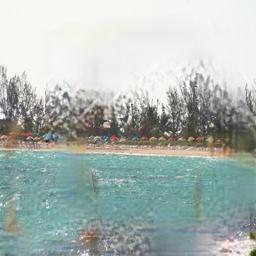}  &
		\includegraphics[width=.16\textwidth]{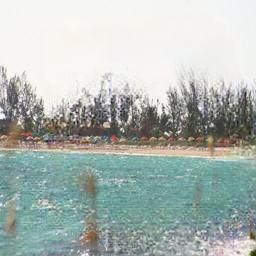}  \\
		
		\includegraphics[width=.16\textwidth]{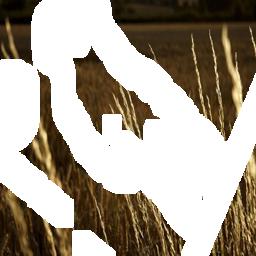}  &
		\includegraphics[width=.16\textwidth]{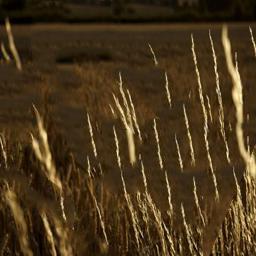}  &
		\includegraphics[width=.16\textwidth]{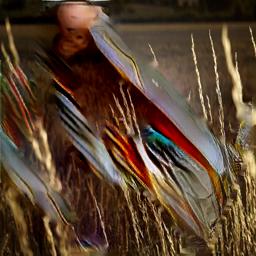}  &
		\includegraphics[width=.16\textwidth]{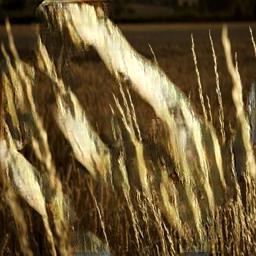}  &
		\includegraphics[width=.16\textwidth]{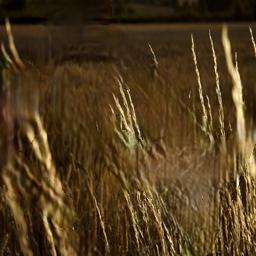}  &
		\includegraphics[width=.16\textwidth]{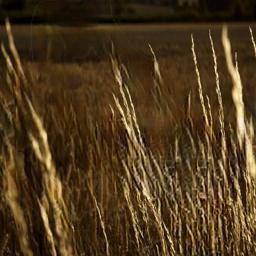}  \\
		
		\includegraphics[width=.16\textwidth]{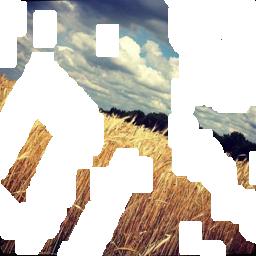}  &
		\includegraphics[width=.16\textwidth]{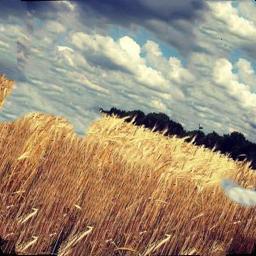}  &
		\includegraphics[width=.16\textwidth]{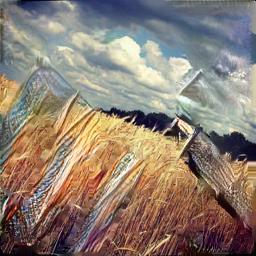}  &
		\includegraphics[width=.16\textwidth]{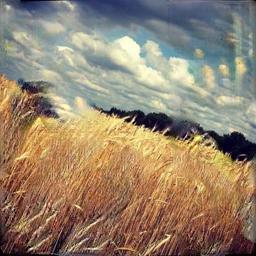}  &
		\includegraphics[width=.16\textwidth]{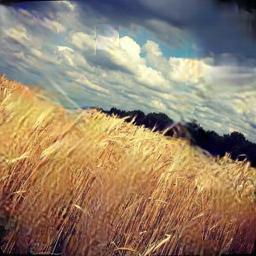}  &
		\includegraphics[width=.16\textwidth]{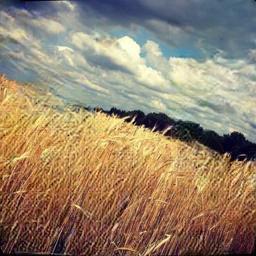}  \\
		
		\includegraphics[width=.16\textwidth]{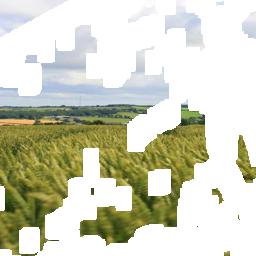}  &
		\includegraphics[width=.16\textwidth]{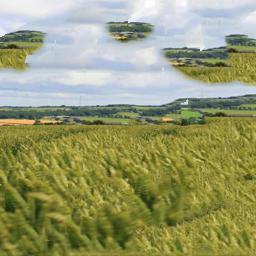}  &
		\includegraphics[width=.16\textwidth]{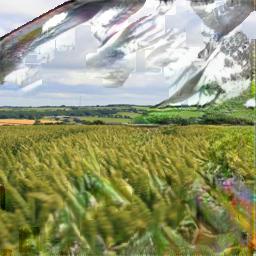}  &
		\includegraphics[width=.16\textwidth]{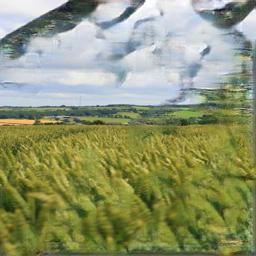}  &
		\includegraphics[width=.16\textwidth]{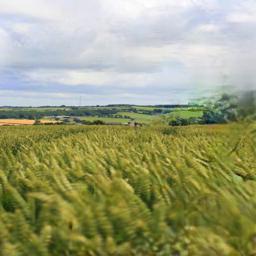}  &
		\includegraphics[width=.16\textwidth]{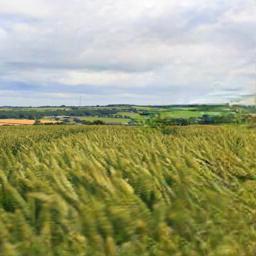}  \\

		Input & PM~\cite{Barnes:2009:PAR} & GL~\cite{IizukaGL} & CA~\cite{yu2018generative} & PConv~\cite{partialconv2017} & Ours \\
		
	\end{tabular}
	\caption{Qualitative comparison on Places dataset. Comparison with PatchMatch (PM)~\cite{Barnes:2009:PAR}, Global\&Local (GL)~\cite{IizukaGL}, Context Attention (CA)~\cite{yu2018generative}, and Partial Convolution (PConv)~\cite{partialconv2017}. All images are scaled to $256 \times 256$.}
	\label{fig:places2}
\end{figure*}

More comparisons with PatchMatch (PM)~\cite{Barnes:2009:PAR}, Global\&Local (GL)~\cite{IizukaGL}, Context Attention (CA)~\cite{yu2018generative}, and Partial Convolution (Pconv)~\cite{partialconv2017} are also conducted. Fig.~\ref{fig:paris1}, ~\ref{fig:paris2} and ~\ref{fig:places2} show the qualitative comparison on Paris StreetView dataset and Places dataset. 
For Paris StreetView~\cite{doersch2015makes} dataset, we use its original splits, $14,900$ images for training, and $100$ images for testing.

For Places~\cite{zhou2017places} dataset, $10$ categories from the total $365$ categories are choosed for training our LBAM model, they are: \textit{apartment\_building\_outdoor}, \textit{beach}, \textit{house}, \textit{ocean}, \textit{sky}, \textit{throne\_room}, \textit{tower}, \textit{tundra}, \textit{valley} and \textit{wheat\_field}. We gather all  $5000$ images of each category to form our training set of $50,000$ images. The validation set from each category of $1,000$ images into two equal non-overlapped sets of $500$ images respectively for validation and testing.
It can be seen that our model performs better in producing both global consistency and fine-detailed structures. 

\subsection*{Object removal on real world images.}
\begin{figure*}[hbt]
	\setlength{\tabcolsep}{2.0pt}
	\centering
	\begin{tabular}{cccccc}
		\includegraphics[width=.16\textwidth]{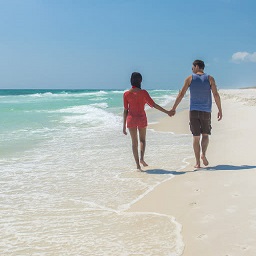}  &
		\includegraphics[width=.16\textwidth]{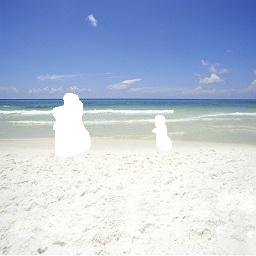}  &
		\includegraphics[width=.16\textwidth]{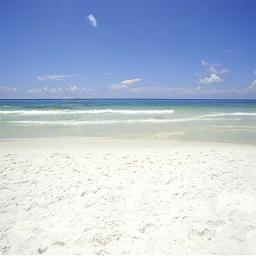}  &
		\includegraphics[width=.16\textwidth]{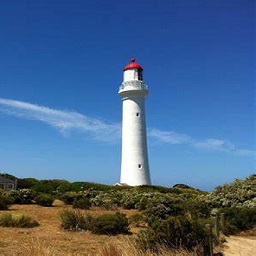}  &
		\includegraphics[width=.16\textwidth]{Object-Removal/input_002}  &
		\includegraphics[width=.16\textwidth]{Object-Removal/ours_002}  \\ 
		
		\includegraphics[width=.16\textwidth]{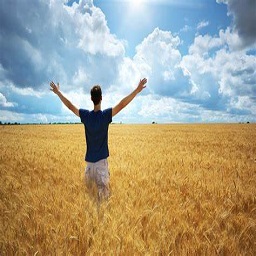}  &
		\includegraphics[width=.16\textwidth]{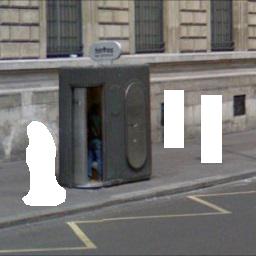}  &
		\includegraphics[width=.16\textwidth]{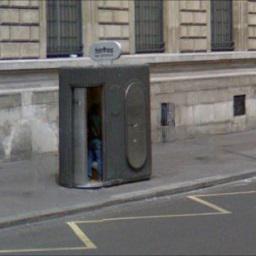}  &
		\includegraphics[width=.16\textwidth]{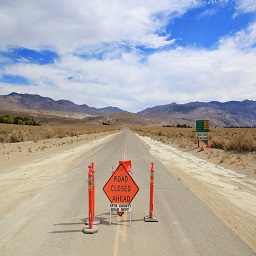}  &
		\includegraphics[width=.16\textwidth]{Object-Removal/input_004}  &
		\includegraphics[width=.16\textwidth]{Object-Removal/ours_004}  \\ 
		
		\includegraphics[width=.16\textwidth]{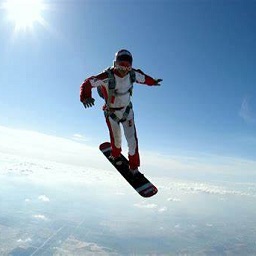}  &
		\includegraphics[width=.16\textwidth]{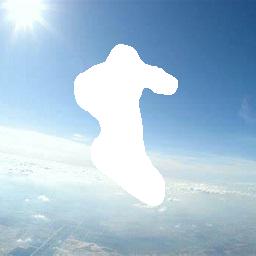}  &
		\includegraphics[width=.16\textwidth]{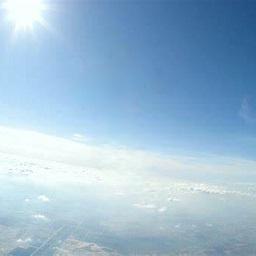}  &
		\includegraphics[width=.16\textwidth]{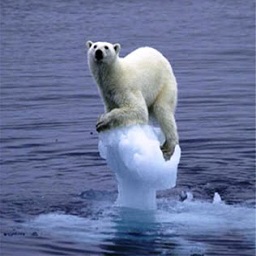}  &
		\includegraphics[width=.16\textwidth]{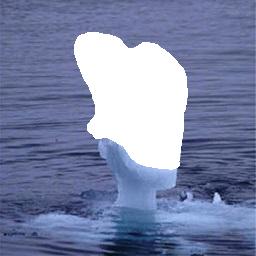}  &
		\includegraphics[width=.16\textwidth]{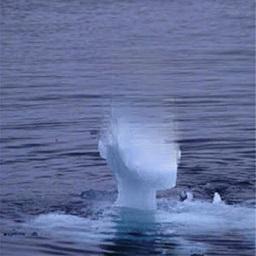}  \\ 
		
		\includegraphics[width=.16\textwidth]{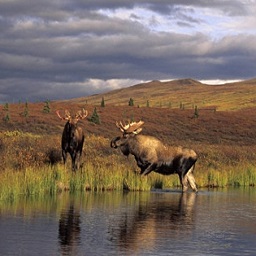}  &
		\includegraphics[width=.16\textwidth]{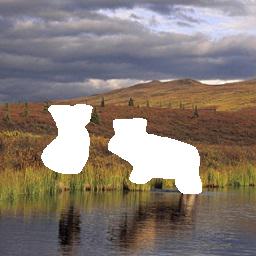}  &
		\includegraphics[width=.16\textwidth]{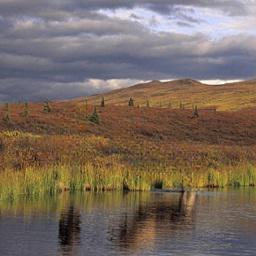}  &
		\includegraphics[width=.16\textwidth]{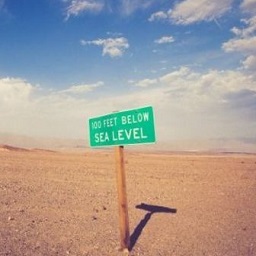}  &
		\includegraphics[width=.16\textwidth]{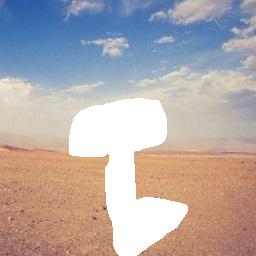}  &
		\includegraphics[width=.16\textwidth]{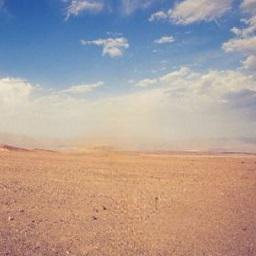}  \\ 
		
		\includegraphics[width=.16\textwidth]{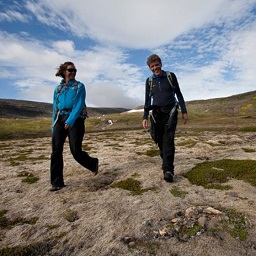}  &
		\includegraphics[width=.16\textwidth]{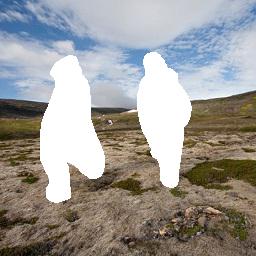}  &
		\includegraphics[width=.16\textwidth]{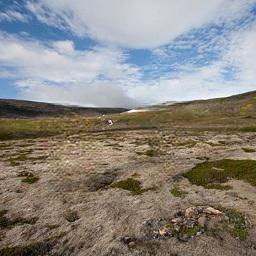}  &
		\includegraphics[width=.16\textwidth]{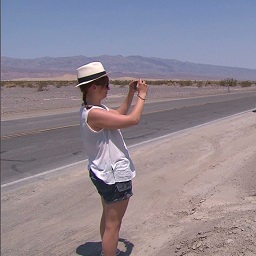}  &
		\includegraphics[width=.16\textwidth]{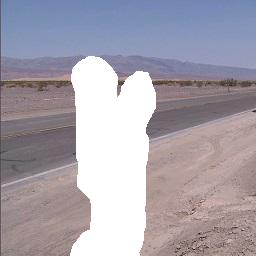}  &
		\includegraphics[width=.16\textwidth]{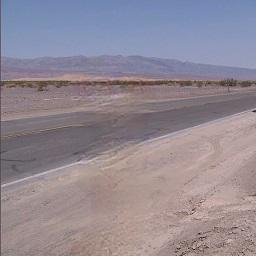}  \\

		\includegraphics[width=.16\textwidth]{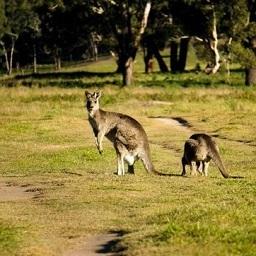}  &
		\includegraphics[width=.16\textwidth]{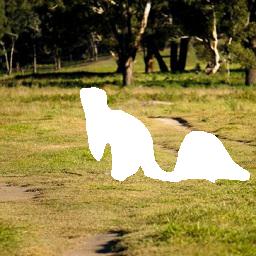}  &
		\includegraphics[width=.16\textwidth]{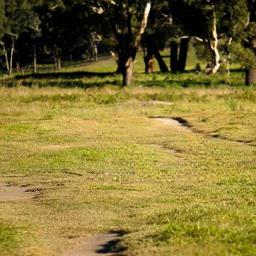}  &
		\includegraphics[width=.16\textwidth]{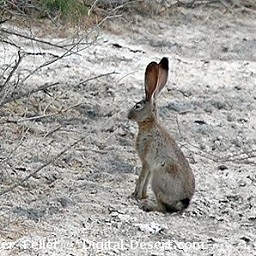}  &
		\includegraphics[width=.16\textwidth]{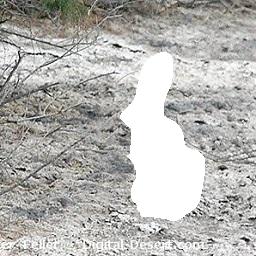}  &
		\includegraphics[width=.16\textwidth]{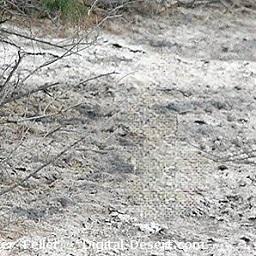}  \\ 
		
		\includegraphics[width=.16\textwidth]{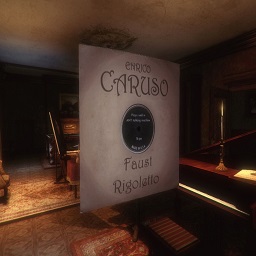}  &
		\includegraphics[width=.16\textwidth]{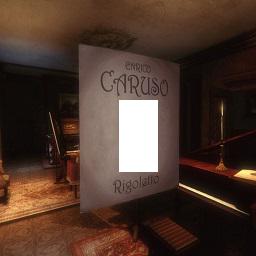}  &
		\includegraphics[width=.16\textwidth]{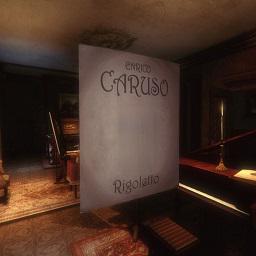}  &
		\includegraphics[width=.16\textwidth]{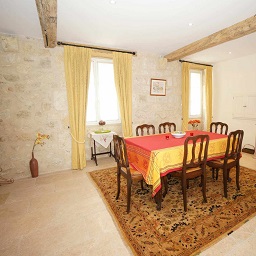}  &
		\includegraphics[width=.16\textwidth]{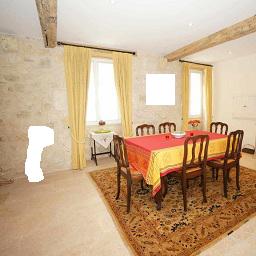}  &
		\includegraphics[width=.16\textwidth]{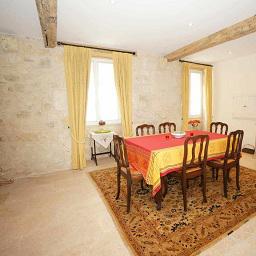}  \\ 
		
		Original Image & Input & Ours & Original Image & Input & Ours \\
		
	\end{tabular}
	\caption{Results of our LBAM on object removal task of real world images. All images are scaled to $256 \times 256$.}
	\label{fig:removal}
\end{figure*}

\vspace{-0.6em}
Finally, we apply our model trained on Places dataset for object removal on real world images. As shown in Fig.~\ref{fig:removal}, although these images contain different objects, background, context and shapes, even some of them have large portion masked regions, our model can handle them well, demonstrating its practicability and generalization ability of our LBAM model.

\begin{table*}[htb]
	\centering
	\caption{The architecture of the discriminator. BN represents BatchNorm, LReLU denotes leaky ReLU with slope of 0.2, and $M$ represents mask with zeros denote the missing pixels and ones denote the remained pixels. }
	
	\begin{tabular}{cccc}
		\hline
		\textbf{Input}: & Image ($256 \times 256 \times 3$)  $ * M$ & \textbf{Input}: & Image ($256 \times 256 \times 3$) $ * (1 - M)$\\ \hline
		[Layer 1-1] & Conv.($4,4,64$), stride = 2; LReLU; & [Layer 1-2]  & Conv.($4,4,64$), stride = 2; LReLU;\\ \hline
		[Layer 2-1] & Conv.($4,4,128$), stride = 2; BN; LReLU; & [Layer 2-2] & Conv.($4,4,128$), stride = 2; BN; LReLU; \\ \hline
		[Layer 3-1] & Conv.($4,4,256$), stride = 2; BN; LReLU; & [Layer 3-2] & Conv.($4,4,256$), stride = 2; BN; LReLU; \\ \hline
		[Layer 4-1] & Conv.($4,4,512$), stride = 2; BN; LReLU; & [Layer 4-2] & Conv.($4,4,512$), stride = 2; BN; LReLU;\\ \hline
		[Layer 5-1] & Conv.($4,4,512$), stride = 2; BN; LReLU;& [Layer 5-2] & Conv.($4,4,512$), stride = 2; BN; LReLU;\\ \hline
		[Layer 6-1] & Conv.($4,4,512$), stride = 2; BN; LReLU;& [Layer 6-2] &  Conv.($4,4,512$), stride = 2; BN; LReLU;\\ \hline
		&        \multicolumn{3}{c}{Concatenate(Layer 6-1, Layer 6-2);}  \\ \hline
		[Layer 7] & \multicolumn{3}{c}{Conv.($4,4,1$), stride = 0; Sigmoid;}  \\ \hline
		
		\textbf{Output}: & \multicolumn{3}{c}{Real or Fake ($1\times 1 \times 1$) } \\
		\hline
	\end{tabular}
	\label{table:discriminator}
\end{table*}

\begin{table*}[htb]
	\centering
	\caption{The architecture of our LBAM model. Ewp() means element-wise product, Cat() represents feature concatenation operation, $g_{A}(\cdot)$ denotes asymmetric Gaussian-shaped form activation function of Eqn. (9), and $g_{M}(\cdot)$ denotes mask updating function of Eqn. (8), BN represents BatchNorm, LReLU denotes leaky ReLU with slope of 0.2,  and $M^{in}$ represents mask with zeros indicating the missing pixels and ones indicating the remained pixels. Note that $g_{A}(\cdot)$ and $g_{M}(\cdot)$ are unique among layers and do not share its parameters. }
	
	\begin{tabular}{cccc}
		\hline
		\multicolumn{2}{c}{Our Modified U-Net} & \multicolumn{2}{c}{Learnable Bidirectional Attention Maps} \\ \hline
		\textbf{Input}: & Image ($256 \times 256 \times 3$) & \textbf{Input}: & $M^{in}$ ($256 \times 256 \times 3$) \\ \hline
		
		[Layer 1-1] & Conv.($4,4,64$), stride = 2; & [Layer 1-2]  & Conv.($4,4,64$), stride = 2; \\ \hline
		& \multicolumn{1}{c}{Ewp(Layer 1-1, $g_A(\mbox{Layer 1-2})$); LReLU;} & \\ \hline
		
		[Layer 2-1] & Conv.($4,4,128$), stride = 2; & [Layer 2-2] & $g_{M}(\cdot) $; Conv.($4,4,128$), stride = 2; \\ \hline
		& \multicolumn{1}{c}{Ewp(Layer 2-1, $g_A(\mbox{Layer 2-2})$); BN; LReLU;} & \\ \hline
		
		[Layer 3-1] & Conv.($4,4,256$), stride = 2; & [Layer 3-2] & $g_{M}(\cdot) $; Conv.($4,4,256$), stride = 2; \\ \hline
		& \multicolumn{1}{c}{Ewp(Layer 3-1, $g_A(\mbox{Layer 3-2})$); BN; LReLU;} & \\ \hline
		
		[Layer 4-1] & Conv.($4,4,512$), stride = 2; & [Layer 4-2] & $g_{M}(\cdot) $; Conv.($4,4,512$), stride = 2;  \\ \hline
		& \multicolumn{1}{c}{Ewp(Layer 4-1, $g_A(\mbox{Layer 4-2})$); BN; LReLU;} & \\ \hline
		
		[Layer 5-1] & Conv.($4,4,512$), stride = 2; & [Layer 5-2] & $g_{M}(\cdot) $; Conv.($4,4,512$), stride = 2; \\ \hline
		& \multicolumn{1}{c}{\small{Ewp(Layer 5-1, $g_A(\mbox{Layer 5-2})$); BN; LReLU;}} & \\ \hline
		
		[Layer 6-1] & Conv.($4,4,512$), stride = 2; & [Layer 6-2] & $g_{M}(\cdot) $; Conv.($4,4,512$), stride = 2; \\ \hline
		& \multicolumn{1}{c}{\small{Ewp(Layer 6-1, $g_A(\mbox{Layer 6-2})$); BN; LReLU;}} & \\ \hline
		
		[Layer 7-1] & Conv.($4,4,512$), stride = 2;& [Layer 7-2] & $g_{M}(\cdot) $; Conv.($4,4,512$), stride = 2; \\ \hline
		& \multicolumn{1}{c}{\small{Ewp(Layer 7-1, $g_A(\mbox{Layer 7-2})$); BN; LReLU;}} & \\ \hline
		
		[Layer 8-1] & DeConv.($4,4,512$), stride = 2;& [Layer 6-3] & $g_{M}(\cdot) $; Conv.($4,4,512$), stride = 2; \\ \hline
		\multicolumn{3}{c}{\small{Ewp(Cat(Layer 8-1, Layer 6-1), Cat($g_A(\mbox{Layer 6-3})$, $g_A(\mbox{Layer 6-2})$));BN; LReLU;}} & \\ \hline
		
		[Layer 9-1] & DeConv.($4,4,512$), stride = 2; &[Layer 5-3] & $g_{M}(\cdot) $; Conv.($4,4,512$), stride = 2; \\ \hline
		\multicolumn{3}{c}{\small{Ewp(Cat(Layer 9-1, Layer 5-1), Cat($g_A(\mbox{Layer 5-3})$, $g_A(\mbox{Layer 5-2})$));BN; LReLU;}} & \\ \hline
		
		[Layer 10-1] & DeConv.($4,4,512$), stride = 2; &[Layer 4-3] & $g_{M}(\cdot) $; Conv.($4,4,512$), stride = 2; \\ \hline
		\multicolumn{3}{c}{\small{Ewp(Cat(Layer 10-1, Layer 4-1), Cat($g_A(\mbox{Layer 4-3})$, $g_A(\mbox{Layer 4-2})$));BN; LReLU;}} & \\ \hline
		
		[Layer 11-1] & DeConv.($4,4,256$), stride = 2; & [Layer 3-3]& $g_{M}(\cdot) $; Conv.($4,4,256$), stride = 2; \\ \hline
		\multicolumn{3}{c}{\small{Ewp(Cat(Layer 11-1, Layer 3-1), Cat($g_A(\mbox{Layer 3-3})$, $g_A(\mbox{Layer 3-2})$));BN; LReLU;}} & \\ \hline
		
		[Layer 12-1] & DeConv.($4,4,128$), stride = 2; & [Layer 2-3]& $g_{M}(\cdot) $; Conv.($4,4,128$), stride = 2; \\ \hline
		\multicolumn{3}{c}{\small{Ewp(Cat(Layer 12-1, Layer 2-1), Cat($g_A(\mbox{Layer 2-3})$, $g_A(\mbox{Layer 2-2})$));BN; LReLU;}} & \\ \hline
		
		[Layer 13-1] & DeConv.($4,4,64$), stride = 2; & [Layer 1-3]& Conv.($4,4,64$), stride = 2; \\ \hline
		\multicolumn{3}{c}{\small{Ewp(Cat(Layer 13-1, Layer 1-1), Cat($g_A(\mbox{Layer 1-3})$, $g_A(\mbox{Layer 1-2})$));BN; LReLU;}} \\ \hline
		
		[Layer 14-1] & DeConv.($4,4,3$), stride = 2; tanh;& \textbf{Input}: & $1 - M^{in}$ ($256 \times 256 \times 3$) \\ \hline
		
		\textbf{Output}: &Final result ($256 \times 256 \times 3$) &  \multicolumn{2}{c}{Reverse Attention Maps} \\ \hline
		
		\hline
	\end{tabular}
	\label{table:generator}
\end{table*}

\end{appendix}

\end{document}